\documentclass[10pt,twocolumn,letterpaper]{article}

\usepackage[pagenumbers]{cvpr} % To force page numbers, e.g. for an arXiv version

 \usepackage{soul}
\usepackage{bm}
\usepackage{tcolorbox}
\usepackage{csquotes}
\usepackage{stfloats}

\definecolor{cvprblue}{rgb}{0.21,0.49,0.74}
\usepackage[pagebackref,breaklinks,colorlinks,allcolors=cvprblue]{hyperref}

\def\paperID{} % *** Enter the Paper ID here; old paper id: 20496; new paper id: 41712
\def\confName{CVPR}
\def\confYear{2026}

\title{A Framework for Evaluating Zero-Shot Image Generation\\in Concept-based Explainability}
\author{
Giacomo Astolfi$^{1}$\thanks{These authors contributed equally.\\ \indent Correspondence to: \texttt{riccardo.campi@polimi.it}.}\\
{\tt\small giacomo.astolfi@mail.polimi.it}
\and
Matteo Bianchi$^{1}$\footnotemark[1]\\
{\tt\small matteo.bianchi@polimi.it}
\and
Riccardo Campi$^{1}$\footnotemark[1]\\
{\tt\small riccardo.campi@polimi.it}
\and
Antonio De Santis$^{1}$\\
{\tt\small antonio.desantis@polimi.it}
\and
Marco Brambilla$^{1}$\\
{\tt\small marco.brambilla@polimi.it}
\and
{$^1$Politecnico di Milano, DEIB, Italy} \quad
}

\begin{document}
\maketitle
\begin{abstract}
Concept-based Explainable Artificial Intelligence (XAI) interprets deep learning models using human-understandable visual features (e.g., textures or object parts) by linking internal representations to class predictions, thereby bridging the gap between low-level image data and high-level semantics.
A major challenge, however, is the reliance on large sets of labeled images to represent each concept, which limits scalability.
In this work, we investigate the use of zero-shot Text-to-Image (T2I) generative models as a source of synthetic concept datasets for concept-based XAI methods. Specifically, we generate concepts using predefined prompts and evaluate their faithfulness to real ones through four complementary analyses: (1) comparing synthetic vs. real concept images via concept representation similarity; (2) evaluating their intra-similarity by comparing pairs of subsets of the same concept with progressively increasing size; (3) evaluating their performance for downstream explanation tasks using relevant class images; (4) evaluating how removing a concept from tested class images affects explanations of generated concepts.
While current T2I generative models promise a shortcut to concept-based XAI, our study highlights challenges and raises open questions about the use of synthetic data generated by zero-shot pipelines in model analyses. The resulting dataset is available at {\small \url{https://github.com/DataSciencePolimi/ZeroShot-T2I-Concepts}}.
\end{abstract}    
\section{Introduction}
\label{sec:intro}

\begin{figure}[t]
  \centering
   \includegraphics[width=0.95\linewidth]{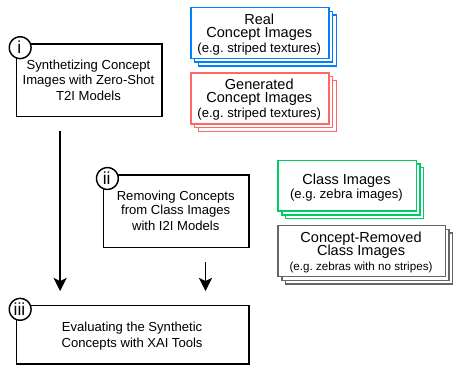}

   \caption{High-level schematic representation of the proposed framework. It encompasses \raisebox{.5pt}{\textcircled{\raisebox{-.5pt} {\footnotesize i}}} synthesis of concept images with zero-shot T2I models and a predefined prompt, \raisebox{.5pt}{\textcircled{\raisebox{-0.5pt} {\footnotesize ii}}} generation of counterfactual class images (i.e., without the concepts) with an I2I model using a fixed prompt, and \raisebox{.5pt}{\textcircled{\raisebox{-0.5pt} {\footnotesize iii}}} evaluation of the synthetic concepts against real ones by comparing both original and ablated class images with XAI tools.}
   \label{fig:mini_pipeline}
\end{figure}

Convolutional Neural Networks (CNNs) and Vision Transformers (ViTs) excel at extracting complex patterns from data, but their decision-making processes are opaque and hard to understand. Explainable AI (XAI) aims to reveal this hidden reasoning, enabling model debugging, bias detection, and trustworthy decisions in critical applications~\cite{arrieta}.
%In computer vision, particularly 
For image classification tasks, many XAI methods link internal activations to human-readable concepts, measuring, for instance, the contribution of a concept to a class's score~\cite{XAI}.
Testing with Concept Activation Vectors (TCAV) \cite{tcav} first introduced this paradigm by representing each concept as a direction in the latent space, known as a Concept Activation Vector (CAV), and by measuring its influence on model predictions. Building on this, Visual-TCAV \cite{vtcav} improves concept magnitude estimation and localizes each concept in an image via a concept map.

These approaches require collecting images that illustrate user-defined semantic concepts, a process that is labor‑intensive and often impractical, limiting both scalability and the variety of concepts that can be studied. To address these limitations, researchers are increasingly relying on synthetic concepts, which enable the (semi-)automated and scalable creation of training data for a much wider range of concepts~\cite{LGCAV}.
However, when Text-to-Image (T2I) models are involved in the synthesis process, the resulting concepts may contain visual artifacts and often lack the richness and variability of real-world data. This can introduce bias, lower the quality of explanations, and amplify spurious model correlations, undermining user trust~\cite{SCG}.

In this work, we present a concept generation framework that %first uses zero-shot T2I generative models based on predefined prompts to produce concept images and then 
investigates how faithfully synthetic concepts match their real counterparts under different perspectives (\Cref{fig:mini_pipeline}).
In particular, this work uses zero-shot T2I generative models based on predefined prompts to produce concept images to address four complementary research questions:

%i have two datasets containing concept images. The first one has ai-generated concept images (the goal of the overall work is to evaluating their quality), while the second one has real concept images. Concepts are the same across the two datasets. For the first research question, i compute the CAV using TCAV and VTCAV of the real and generated concepts. If i compare the two cavs, for each concept, what i'm measuring? For the second research question, I'm using these concept images for a downstream task, such as computing the concept attribution for relevant classes with a user-defined concept-based XAI method, evaluating how much the results change. A third experiment takes these relevant class images, removes the concepts from them using a prompt with an image-to-image model, and evaluates the quality of the removal by examining whether the classification performance remains stable before and after the removal. A fourth experiment utilizes the class images after removal to evaluate concept attribution for both generated and real concepts. Write the four research questions

\begin{itemize}
    %\item \textbf{Representational alignment}: Do synthetic concepts produced by text‑to‑image models activate neural representations that are comparable to those evoked by real‑world concept images?
    \item[RQ1] \textbf{Alignment of Concept Representations}: To what extent do the concept representations (i.e., CAVs) derived from synthetic images align with the ones derived from real images?
    
    \item[RQ2] \textbf{Concept Representation Intra-Similarity}: How does, for each concept, increasing the number of concept representation training images affect their internal pairwise similarity, and how does this effect differ between real‑image and synthetic‑image sets?
    
    \item[RQ3] \textbf{Impact on Downstream Explanation Task}: How does substituting real concept images with synthetic ones affect explanations (e.g., concept attribution scores) produced for target classes?
    
    \item[RQ4] \textbf{Counterfactual Testing on Downstream Task}: If a concept is removed from class images using a prompt‑driven Image‑to‑Image (I2I) model, how do explanations change for both real and generated concept images?
\end{itemize}

%\begin{figure*}
%  \centering
%  \begin{subfigure}{0.68\linewidth}
%    \fbox{\rule{0pt}{2in} \rule{.9\linewidth}{0pt}}
%    \caption{An example of a subfigure.}
%    \label{fig:short-a}
%  \end{subfigure}
%  \hfill
% \begin{subfigure}{0.28\linewidth}
%    \fbox{\rule{0pt}{2in} \rule{.9\linewidth}{0pt}}
%    \caption{Another example of a subfigure.}
%    \label{fig:short-b}
%  \end{subfigure}
%  \caption{Example of a short caption, which should be centered.}
%  \label{fig:short}
%\end{figure*}

\section{Related Work}
\label{sec:related}

XAI has gained significant attention for its goal of making machine learning models more interpretable and understandable to humans \citep{arrieta}. This is especially important for neural networks, whose internal operations are often hard to interpret. %In the field of computer vision, gradient-based techniques generate saliency maps to indicate which image regions influence predictions~\cite{GRADCAM,IG}. However, these maps highlight where the model looks, not what it sees or why those regions matter~\cite{no_saliency_maps}.

In computer vision, to understand which high-level features influence model decisions, researchers have explored concept-based explanation methods, which assess the impact of human-interpretable concepts, ranging from shapes and colors to textures, object parts, and abstract traits like happiness or sadness, on a model's predictions~\cite{molnar2022,XAI}.

A widely adopted technique, mostly used on convolutional architectures, is TCAV~\cite{tcav}, which measures a model's sensitivity to user-defined concepts during image classification. %\hl{This approach, in contrast to unsupervised methods that extract concepts automatically and do not allow the user to control selection}~\cite{ACE,ICE,craft}, enables the explicit testing of specific patterns defined by the analyst, which is the scenario this paper focuses on.
To produce a CAV, an analyst collects example images representing a concept along with negative examples (e.g., random images). A linear classifier is trained to distinguish its activations within a specific layer, and the CAV is defined as the vector orthogonal to the classifier's decision boundary. TCAV then compares this vector with the gradient of a target class in a set of test images to determine whether the concept positively or negatively influences the output.
Later work showed that CAVs can also be computed directly via the Difference of Means (DoM) method, improving efficiency and stability to small perturbations~\cite{MartinPhd,patterncav}.

Visual-TCAV further improved the approach by replacing the flattening operation with a Global Average Pooling (GAP) during centroid computation, producing pooled-CAVs that better separate spatial features from conceptual representations~\cite{vtcav}. Additionally, its concept attribution score measures the magnitude of a concept’s influence on the model’s prediction, not just its sensitivity to that concept. It is computed by aggregating normalized integrated gradient attributions across feature maps using the normalized pooled-CAV weights, producing a spatial attribution map, which is subsequently masked by the normalized concept map and summed over spatial locations to obtain a scalar attribution score for the target class.

CAVs built from user-defined concepts %effectively detect biases 
%, such as ethnicity, 
% but 
require manually curated image datasets for accurate detection, which can be time-consuming.
Text2Concept~\cite{text2concept} addresses this problem by utilizing CLIP~\cite{clip} to generate CAVs from text descriptions, and training a linear layer to map CLIP features to a CNN layer. Similarly, Language-Guided CAV (LG-CAV) builds on this idea, proposing the use of CLIP, along with additional modules, to enhance concept quality and train CAVs from textual descriptions.~\cite{LGCAV}.
%However, this may reduce interpretability and explanation quality, as it fully depends on a black-box model that analysts cannot visually validate against the intended concept.
%\hl{On the other hand, an emerging line of work leverages T2I generation models to automate the production of images that serve as concept or feature sets.} %This approach primarily aims to reduce the need for analysts to manually curate concept images.
%Some works employed them to identify or modify the minimal set of features necessary to alter a classifier’s prediction for a given image, a process known as Counterfactual Explanation (CE)~\cite{TIME,counterfactuals_mahalanobis}, although this may introduce artifacts or deviate from the true data manifold, thereby reducing their explanatory reliability.
%A notable work is the Multimodal Automated Interpretability Agent (MAIA), which incorporates a pre-trained vision-language model into its methodology to automate model understanding tasks, such as generating neuron-level feature descriptions~\cite{maia}. Nevertheless, its scope differs somewhat from the objectives pursued here.
On the other hand, an emerging line of work explored the alignment between real and synthetic CAVs generated via zero-shot and fine-tuned T2I models, enabling the refinement of generated images using a small set of real examples~\cite{SCG}. However, it revealed limitations, such as drops in semantic alignment with complex concepts (e.g., 3D objects) or the presence of artifacts.

Building on prior efforts, this work introduces a concept-generation and validation pipeline that leverages zero-shot T2I models and predefined generation prompts. The goal is to assess how faithfully synthetic concepts replicate their real-image counterparts, both in terms of activation space representations and their influence on downstream explanation tasks.

\section{Method}
\label{sec:method}

\begin{figure}[t]
  \centering
   \includegraphics[width=1\linewidth]{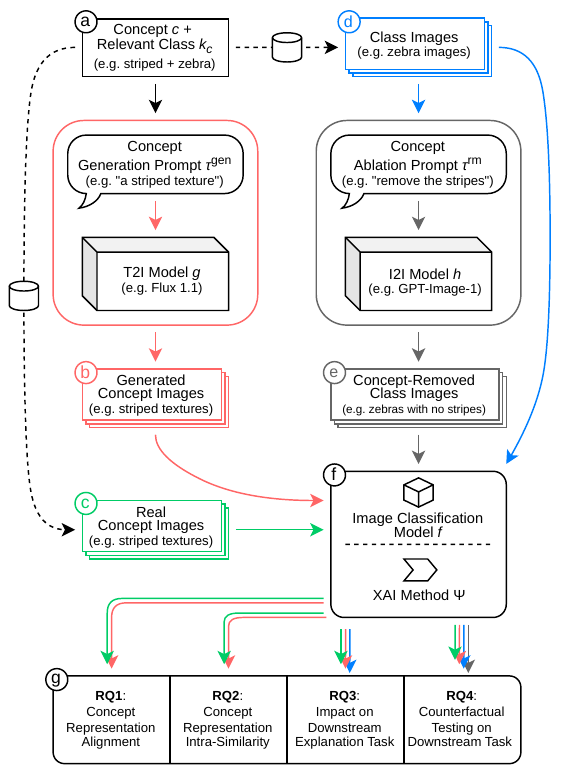}

   \caption{Pipeline for generating synthetic concepts using zero-shot T2I models and predefined prompts, and evaluating them for their use in concept-based, user-defined XAI methods. \raisebox{.5pt}{\textcircled{\raisebox{-.5pt} {\footnotesize a}}} A concept $c$ and its appropriate class $k_c$ are selected, then \raisebox{.5pt}{\textcircled{\raisebox{-1pt} {\footnotesize b}}} synthetic concept images are generated from prompt templates $\tau^\text{gen}$ using a T2I model $g$, while \raisebox{.5pt}{\textcircled{\raisebox{.5pt} {\footnotesize c}}} real concept images are collected from data. \raisebox{.5pt}{\textcircled{\raisebox{-.5pt} {\footnotesize d}}} Class $k_c$ images containing the concept are gathered and then \raisebox{.5pt}{\textcircled{\raisebox{.5pt} {\footnotesize e}}} edited using I2I models $h$ and prompt templates $\tau^\text{rm}$ to remove $c$, resulting in concept-removed class images. \raisebox{.5pt}{\textcircled{\raisebox{-1pt} {\footnotesize f}}} An image classification model $f$ and an XAI method $\Psi$ are used on both synthetic and real images to compute CAVs. \raisebox{.5pt}{\textcircled{\raisebox{.5pt} {\footnotesize g}}} Four research questions are addressed: \textbf{RQ1} the alignment between synthetic and real representations, \textbf{RQ2} their intra-similarity as the number of images grows, \textbf{RQ3} the impact on downstream explanation tasks, and \textbf{RQ4} the consistency of explanations after concept removal.}
   \label{fig:pipeline}
\end{figure}

The idea behind this work is to evaluate the faithfulness of synthetic concepts to real ones when integrating zero-shot T2I generative models into a concept-generation workflow for user-defined explainability methods.

\subsection{Synthetic Image T2I Generation}

The proposed framework (\cref{fig:pipeline}) begins by taking in a set of concepts, where each concept $c$ is associated with a type $t_c$ (e.g., texture, object) and a predefined relevant class $k_c$.

Each unique type $t_c$ is associated with a dedicated prompt template $\tau_{t_c}^{\text{gen}}(c)$, which takes the concept as input and produces as output its corresponding generation prompt.
These are then passed to a randomly seeded T2I model $g$, which generates a collection of $n_c$ synthetic images $I_c^{\text{gen}}$ for each concept $c$.
For comparison, an equivalent, per-concept set of $m_c$ real images $I_c^{\text{real}}$ is collected.

\subsection{Concept Representation Alignment}

The purpose of evaluating the alignment of concept representations between synthetic and generated data is to obtain an initial quantitative indication of their similarity.

We extract latent representations from concept images using an image classification model $f$ (e.g., a pre-trained CNN), which maps an image to its corresponding latent vector of activations up to a specific layer/block.
Given a fixed concept $c$, generator $g$, and model $f$, we name $A_c^{\text{gen}}$ the resulting synthetic latent activations.
Analogously, $A_c^{\text{real}}$ are the latent activations derived from real concept images.
A set of negative examples $I^{\text{neg}}$ yielding to $A^{\text{neg}}$ is also required.

Subsequently, a user-defined, concept-based XAI method for vision, $\Psi$, is used to compute CAVs via the function $\psi^{\text{cav}}$.
%We introduce now $\Psi$, a user-defined, concept-based XAI method for vision compatible with $f$, which computes a CAV through a function $\psi^{\text{cav}}$.
Specifically, given the concept $c$ and the activations $A_c^{\text{gen}}$, $A_c^{\text{real}}$, and $A^{\text{neg}}$ in the feature space of $f$, it produces their CAVs as
\[
\mathbf{v}_c^{\text{gen}}=\psi^{\text{cav}}(A_c^{\text{gen}}, A^{\text{neg}}), \qquad
\mathbf{v}_c^{\text{real}}=\psi^{\text{cav}}(A_c^{\text{real}}, A^{\text{neg}}) .
\]
To quantify alignment, we compute the pairwise similarity $\rho_c$ using the cosine similarity function $\mathcal{S}$ between the concept representations, which is suited for highlighting perceptually meaningful variations~\cite{perceptual_similarity}:
\[
\boxed{\rho_c=\mathcal{S}( \mathbf{v}_c^{\text{gen}}, \mathbf{v}_c^{\text{real}} )} .
\]
Higher $\rho_c$ indicates stronger alignment between synthetic and real CAVs, while lower values suggest limited semantic overlap.
%$\rho_c$ is expected to reveal partial alignment between synthetic and real CAVs, suggesting limited semantic overlap.

\subsection{Concept Representation Intra-Similarity}

The concept representation intra‑similarity measures how consistent each concept’s CAV is as we increase the number of images used to compute it. Each comparison is performed between two randomly sampled, equal-sized subsets drawn from the same concept. This will enable us to quantify each concept's representational stability as a function of its image set size, both for synthetic and real images.
By examining how these curves differ, we can assess whether synthetic images provide a faithful approximation of real concept variability.

To do so, within each concept $c$, we compute $\boxed{\rho_c^{\text{gen}}(u)}$ as the similarity between pairs of CAVs obtained from equally-sized random subsets of $u$ synthetic images, with $u\leq n_c/2$. The factor of $1/2$ is a constraint that ensures that the two subsets do not overlap.
The same is done for $\boxed{\rho_c^{\text{real}}(u)}$, with $u\leq m_c/2$.

Although $\rho_c^{\text{gen}}(u)$ and $\rho_c^{\text{real}}(u)$ are expected to grow toward 1 with larger $u$~\cite{SCG}, discrepancies at the same subset size indicate that synthetic images may fall short of representing the complexity of real concepts.
%However, if the curves diverge substantially, this indicates that synthetic images do not capture the diversity and complexity of real examples and therefore may not support the same level of concept representation.
%Increasing the number of training images is expected to enhance intra‑similarity across both datasets. However, synthetic images are likely to exhibit a steeper rise than real ones, as they tend to be overly similar to one another. This lack of diversity can limit the capacity to integrate multiple semantic features and to capture subtle nuances and richer relationships representing real, complex concepts.

\subsection{Impact on Downstream Explanation Task}

To evaluate the impact of synthetic concepts on a downstream explanation task, we measure the extent to which these contribute to classifier $f$'s decisions, providing a basis for comparison with real ones.
For each concept $c$ and its relevant target class $k_c$, we evaluate the effectiveness of $\mathbf{v}_c^{\text{gen}}$ by quantifying the influence of $c$ on predicting $k_c$. We rely on $\psi^\text{score}$, which is the importance function defined within $\Psi$.
The evaluation is grounded in activations $A_{k_c}$ derived from class-specific images $I_{k_c}$:
\[
s_c^{\text{gen}}=\psi^{\text{score}}(\mathbf{v}_c^{\text{gen}},k_c,A_{k_c}), \quad
s_c^{\text{real}}=\psi^{\text{score}}(\mathbf{v}_c^{\text{real}},k_c,A_{k_c}).
\]
%The corresponding real-CAV score is denoted $a_c^{\text{real}}$.

We then measure the difference between these scores to evaluate how closely synthetic concepts match the explanatory power of real ones:
\[
\boxed{
\Delta s_c=|s_c^{\text{gen}} - s_c^{\text{real}}| 
} .
\]
Large values of $\Delta s_c$ indicate that the synthetic concept does not reproduce the importance pattern of its real counterpart, signaling a potential mismatch in semantic fidelity.
%If confirmed, results would indicate that synthetic concepts consistently yield different importance scores than real ones, suggesting a gap in semantic fidelity.

\subsection{Counterfactual Testing on Downstream Task}

To further test the influence of a synthetic concept $c$ for explanations, we remove it from class images $I_{k_c}$ and observe the impact on both real and generated concept images in the downstream task before and after concept removal.

The removal process is carried out by an I2I generative model $h$, which is guided by a predefined prompt template $\tau^{\text{rm}}(c)$ to edit each image.
This produces a new set of class images $I_{k_c}^{\text{rm}}$ where the concept is absent.

\subsubsection{Concept Importance Variation After Removal}

Comparing the variation in importance scores across explanations for real and generated images after removal reveals how closely synthetic concepts mirror the explanatory role of real concepts in downstream tasks, as seen through a counterfactual lens. 

First, we compute $A_{k_c}^{\text{rm}}$ as the activations obtained after removing $c$ from class images. Then, we employ them to obtain the synthetic importance score 
\[
s_c^{\text{gen},\text{rm}}=\psi^{\text{score}}(\mathbf{v}_c^{\text{gen}},k_c,A_{k_c}^{\text{rm}}) .
\]
The same process is applied to real images, yielding $s_c^{\text{real},\text{rm}}$.

The importance variations after removal for synthetic and real concept images are then defined as
\[
\boxed{
\Delta s_c^{\text{gen},\text{rm}}=s_c^{\text{gen}} - s_c^{\text{gen},\text{rm}}
} ,
\]
\[
\boxed{
\Delta s_c^{\text{real},\text{rm}}=s_c^{\text{real}} - s_c^{\text{real},\text{rm}}
} .
\]
If synthetic concepts exhibit importance variations similar to those of their real counterparts, this would suggest that they encode comparable semantic information and can be reliably employed in explanation pipelines. Conversely, it indicates that the synthetic concepts fail to capture the same underlying semantics.

\subsubsection{Concept Removal Validation via Accuracy Drop}

%To strengthen the concept importance variation test results evaluating the T2I-generated concepts,
To evaluate the I2I removal process itself, we compute the classification probability of the neural classifier $f$ for class ${k_c}$, independently of the selected layer, on both the original class images $I_{k_c}$ and on the images after removal $I_{k_c}^{\text{rm}}$. We then measure the classification probability drop on $k_c$ as
\[
\boxed{
\Delta P_{k_c}=f_{k_c}(I_{k_c})-f_{k_c}(I_{k_c}^{\text{rm}}) 
} .
\]
%For the experiment, we reuse $\Delta a_c^{\text{real},\text{rm}}$, which represents the importance variation computed using the real concept images before and after concept removal from the class images.

If changes in $\Delta P_{k_c}$ and $\Delta s_c^{\text{real},\text{rm}}$ (i.e., the importance variation after concept removal for real concept images) are positively correlated, then removing the concept from the class-specific images using the I2I model would be considered successful, and the results of counterfactual testing on the downstream task with synthetic images would be stronger.

\section{Experiments}
\label{sec:experiments}

The experimental design investigates how diverse generative models, visual domains, and model architectures impact the faithfulness of synthetic concept images to their real counterparts, introducing several experimental choices that represent unique firsts in this research field.
Our code can be found at {\small \url{https://github.com/DataSciencePolimi/ZeroShot-T2I-Concepts}}.

\subsection{Concepts and Datasets}

We select multiple concept datasets, spanning different levels of visual abstraction, from low-level textures and materials to high-level object semantics.

We include 13 concepts from the Describable Textures Dataset (DTD)~\cite{DTD}, which provides perceptual texture concepts, such as cracked or honeycombed, to test whether synthetic ones capture low-level visual patterns.
Then, we incorporate as concepts a subset of 10 ImageNet~\cite{imagenet} classes, each assigned the type $t_c = \text{object}$, requiring the network to combine multiple conceptual cues (e.g., tattoo, feline).
We also utilize 10 concepts from the Flickr Material Database (FMD)~\cite{FMD}, each with $t_c = \text{texture}$ and comprising both natural and artificial materials, such as foliage and metal. Unlike DTD, it offers greater variability and context, as materials appear under diverse lighting and scene conditions.
Finally, in line with prior studies, we incorporate 8 concepts from search engines (e.g., leopard print, sphere) by manually collecting example images from publicly available sources, ensuring a diverse range of visual samples, whose type can be both object or texture.

Our concept set contains 41 concepts and 3,860 images, with the number of images per concept $m_c$ varying across datasets: 120 for DTD, 50 for ImageNet, 100 for FMD, and 100 for the search engine's concepts. For each concept, we associate a corresponding ImageNet class (e.g., dotted$\rightarrow$dalmatian, chequered$\rightarrow$crossword puzzle), with each class containing 50 images, resulting in a total of 2,100 class images. A full listing of all the concepts used can be found in \Cref{sec:app_concepts}.

\subsection{Experimental Setup}

%To rigorously assess whether synthetic concepts match their real counterparts, our pipeline leverages multiple T2I generative models, feature extraction models, and concept-based XAI methods.

\paragraph{T2I and I2I Generative Models}

The T2I models, denoted by $g$ in out pipeline, are Flux 1.1 Schnell~\cite{flux}, Stable Diffusion 3.5 Medium (SD 3.5)~\cite{sd3}, and GPT-Image-1~\cite{flux,sd3}. These represent distinct paradigms of visual synthesis. Flux employs a transformer-based rectified flow architecture capable of fine-grained spatial reasoning, producing highly detailed images with strong compositional structure. SD 3.5 performs generation in a latent space via a variational autoencoder, achieving high perceptual fidelity and efficiency. GPT-Image-1, by contrast, operates as a multimodal language model that integrates linguistic and visual reasoning, allowing precise adherence to textual instructions but often producing concept images with recurring styles. GPT-Image-1 is also used as an I2I model in our pipeline, $h$, specifically for concept removal from class images via prompt-guided editing.

\paragraph{Image Classification Models}

Models $f$ with heterogeneous internal architectures are essential to ensure the validity of our findings does not depend on a single network design. %for examining whether synthetic concepts exhibit consistent behavior across different levels of representation, ensuring that 
For this reason, we use the last two convolutional layers from four selected architectures as the analytical backbone of our experiments. VGG-16 (\textit{block5\_conv2} and \textit{block5\_conv3}) \cite{vgg}, ResNet-50-V2 ({\textit{conv5\_block2\_out}} and {\textit{conv5\_block3\_out}}) \cite{resnetv2}, and Inception-V3 ({\textit{mixed9}} and {\textit{mixed10}}) \cite{inceptionv3} serve as canonical CNNs, while ConvNeXt ({\textit{add\_34}} and {\textit{add\_35}}) \cite{convnext} has a new-generation convolutional design informed by transformer architecture principles. All networks are pretrained on ImageNet~\cite{imagenet}, which makes them well-suited as feature extractors for general-domain concepts, as in our case.

\paragraph{XAI Methods}

In this work, we use TCAV~\cite{tcav} with DoM and Visual-TCAV~\cite{vtcav} as the main XAI methods $\Psi$ to compute CAVs and importance scores. %Unlike other user-defined, concept-based techniques that primarily assess a network’s sensitivity to a concept, it explicitly quantifies its magnitude on model predictions, making a difference when evaluating the explanatory relevance of concept representations in downstream tasks. We also report results obtained using TCAV~\cite{tcav} with DoM to further validate our main findings.
Additionally, we use CLIP~\cite{clip} as an external validation for the alignment and intra-similarity of concept representations, providing further validation of the results.

\subsection{Concept Generation and Removal}

\begin{figure}[t]
    \centering
    \subfloat[Real]{%
        \includegraphics[width=0.23\columnwidth]{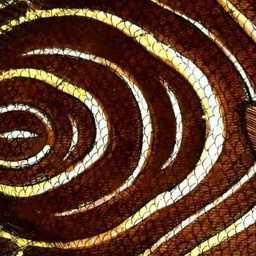}
        \label{fig:striped_real}
    }
    \subfloat[Flux 1.1]{%
        \includegraphics[width=0.23\columnwidth]{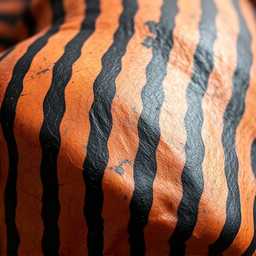}
        \label{fig:striped_flux}
    }
    \subfloat[SD 3.5]{%
        \includegraphics[width=0.23\columnwidth]{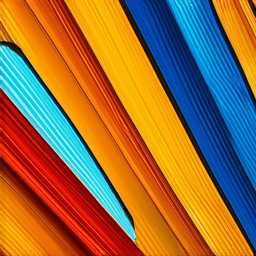}
        \label{fig:striped_sd35}
    }
    \subfloat[GPT-Image-1]{%
        \includegraphics[width=0.23\columnwidth]{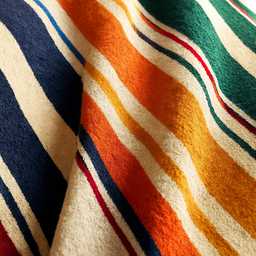}
        \label{fig:striped_gpt1}
    }
    \caption{Example of striped concept images: comparison between a real one and synthetic renditions generated by Flux 1.1 (\ref{fig:striped_flux}), SD 3.5 (\ref{fig:striped_sd35}), and GPT‑Image‑1 (\ref{fig:striped_gpt1}).}
    \label{fig:striped}
\end{figure}

Each synthetic concept $c$ consists of $n_c=200$ images per generative model $g$, which is sufficient to robustly estimate CAVs. All images are preprocessed to each model’s input dimensions, without extra augmentations. \Cref{fig:striped} shows an example of a striped concept, while \Cref{sec:app_gen_concepts} shows a representative subset of the generated concepts.

Concept generation prompts are designed to align with each unique semantic type $t_c$, whether a texture or an object. In particular, each concept type is linked to its prompt template $\tau_{t_c}^{\text{gen}}(c)$, into which the concept name $c$ is dynamically inserted as a parameter: %, avoiding manual bias or prompt tuning:

\begin{tcolorbox}[colback=black!5!white, colframe=black, title=T2I Object Generation Prompt Template] %\small
\textit{A realistic photo of a [CONCEPT] shown clearly in action and in its natural setting. The image should focus on the object's shape, material, and details, captured with natural lighting and minimal background distractions. Photographed from different angles or in natural settings, but always with the object clearly visible and realistically rendered.}
\end{tcolorbox}

\begin{tcolorbox}[colback=black!5!white, colframe=black, title=T2I Texture Generation Prompt Template] %\small
\textit{A realistic close-up of the concept [CONCEPT] as the main subject, shown in high detail and natural lighting. The concept should fill the frame or dominate the image, with clear texture and form. Multiple perspectives and realistic, rich, colorful surfaces are encouraged}.
\end{tcolorbox}

In contrast, the template $\tau^{rm}(c)$ used to remove a concept $c$ from its class images does not depend on its type:

\begin{tcolorbox}[title=I2I Concept Removal Prompt Template,colback=gray!5,colframe=black] %\small
\textit{Completely remove the concept of [CONCEPT] from the image, keeping it as visually similar as possible without [CONCEPT] being present at all.}
\end{tcolorbox}

\begin{figure}[t]
    \centering
    \subfloat[Original]{%
        \includegraphics[width=0.23\columnwidth]{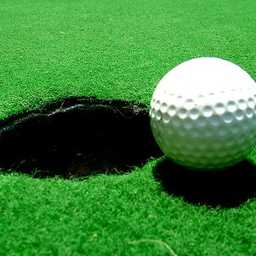}
        \label{fig:golf_ball_original}
    }
    \subfloat[No grass]{%
        \includegraphics[width=0.23\columnwidth]{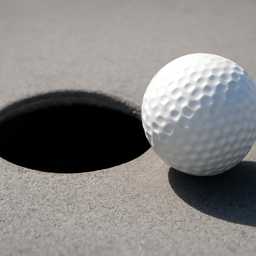}
        \label{fig:golf_ball_ungrass}
    }
    \subfloat[No sphere]{%
        \includegraphics[width=0.23\columnwidth]{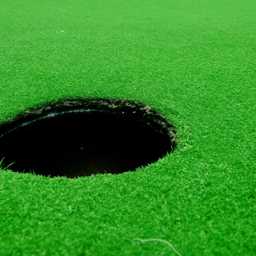}
        \label{fig:golf_ball_unsphere}
    }
    \subfloat[No plastic]{%
        \includegraphics[width=0.23\columnwidth]{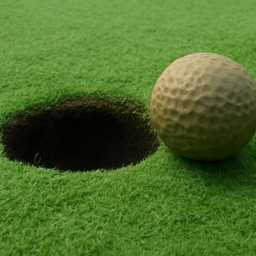}
        \label{fig:golf_ball_unplastic}
    }
    \caption{Example of concept removal from a golf ball class image, eliminating grass (\ref{fig:golf_ball_ungrass}), sphericity (\ref{fig:golf_ball_unsphere}), and plastic (\ref{fig:golf_ball_unplastic}).}
    \label{fig:golf_ball}
\end{figure}

The resulting prompt and an I2I model $h$ along with $I_{k_c}$
%is then passed to the I2I model $h$ along with $I_{k_c}$,
produce a dataset of ablated class images $I_{k_c}^{\text{rm}}$ of the same size as the original ones. An example showing a golf ball image after concept removal is shown in \cref{fig:golf_ball}, while other ablation examples are reported in \cref{sec:app_rem_images}.

%A complete overview of the prompt templates is provided in \Cref{sec:app_prompts}.

%To ensure the robustness of our results, 
Computations involving the generated concepts are carried out with 5 bootstrap replicates, where resampling is performed with replacement from their associated concept image sets at a unitary sampling ratio. At each iteration, a resampled dataset of equal size is constructed and used to recompute both $\mathbf{v}_c^{\text{gen}}$ and the importance scores $s_c^{\text{gen}}$.

\subsection{Hardware and Execution Time}

All local computations took place on a 10 GB NVIDIA RTX 3080 GPU, a 16-core AMD 5800X processor, and 64 GB of system memory. Each concept dataset required 5–10 minutes to synthesize on our hardware, resulting in an average total generation time of 12 hours across both Flux and SD models. 
The remaining concept images, together with those used for concept removal, were generated using GPT‑Image‑1 via the OpenAI API and completed within 8 hours, at a cost of approximately \$150. Subsequent stages, such as feature extraction, CAV computation, and concept importances, were executed locally in 4-6 hours for all architectures and datasets.

\section{Results and Discussion}
\label{sec:results}

Below, we present the results, mirroring the structure of \cref{sec:method}, and discuss each step in detail to address the RQs. 
All results are computed using Visual-TCAV and presented aggregated by layer and by convolutional architecture, with more details in \Cref{sec:app_cos_sim_vtcav,sec:app_importances_vtcav}. \Cref{sec:app_cos_sim_tcav,sec:app_importances_tcav} contain instead TCAV-specific results.

\subsection{Concept Representation Alignment}
\label{subsec:concept_representation_alignment}

\begin{figure}[t]
  \centering
   \includegraphics[width=1\linewidth]{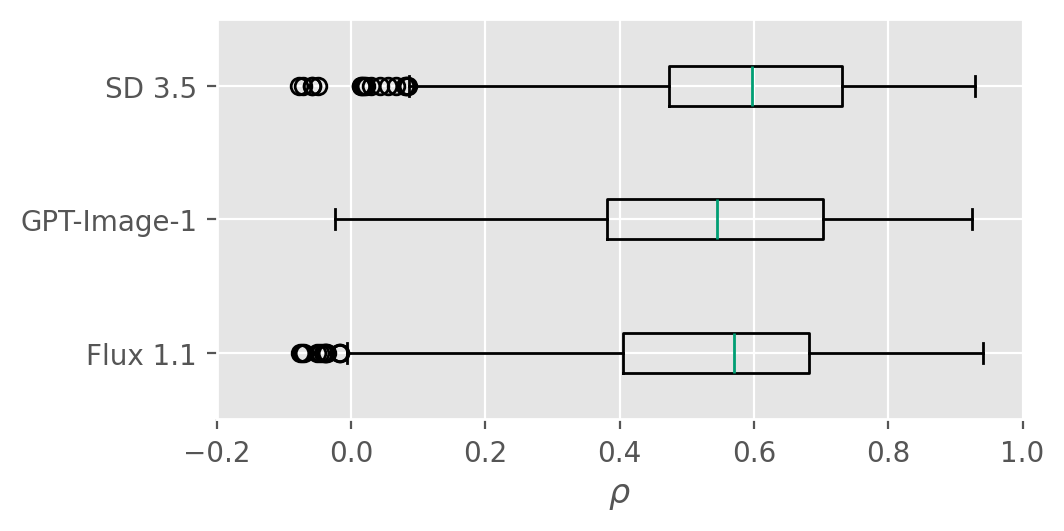}

   \caption{Boxplot showing the distribution of similarities $\rho$ between generated and real concept vectors ($\mathbf{v}_c^{\text{gen}}$ and $\mathbf{v}_c^{\text{real}}$) across different T2I generative models computed using Visual-TCAV. The average similarity scores for Flux 1.1 ($0.541\pm0.204$), GPT-Image-1 ($0.539\pm0.199$), and Stable Diffusion 3.5 ($0.588\pm0.190$) indicate partial alignment between synthetic and real concepts, with enough divergence to suggest semantic differences.}
   \label{fig:cav_similarity}
\end{figure}

After computing $\mathbf{v}_c^{\text{gen}}$ and $\mathbf{v}_c^{\text{real}}$ for each concept $c$, we compute their cosine similarity $\rho_c$ to evaluate their alignment. \Cref{fig:cav_similarity} compares the results from the selected T2I models. Quantitatively, we obtain an average similarity $\bar{\rho}$ over all concepts of $0.541\pm0.204$ for Flux 1.1, $0.539\pm0.199$ for GPT-I-1, and $0.588\pm0.19$ for SD 3.5. More details are presented in \Cref{sec:app_cos_sim_vtcav,sec:app_cos_sim_tcav}.

These results show that real and synthetic CAVs share some similarity but are also sufficiently distinct. Specifically, similarities remain relatively low, and their standard deviations are significant. Therefore, with respect to RQ1, synthetic images can only provide partial alignment with real ones, yielding concepts that may differ. 

To further validate our results, we compute $\rho_c^\text{CLIP}$ by comparing the CLIP embeddings of $I_c^{\text{gen}}$ and $I_c^{\text{real}}$, yielding on average $0.673 \pm 0.097$ for Flux 1.1, $0.706 \pm 0.094$ for GPT-I-1, and $0.708 \pm 0.091$ for SD 3.5, The pattern agrees with the one presented above, with sufficient divergence between real and synthetic concepts.

\subsection{Concept Representation Intra-Similarity}

\begin{figure}[t]
  \centering
   \includegraphics[width=1\linewidth]{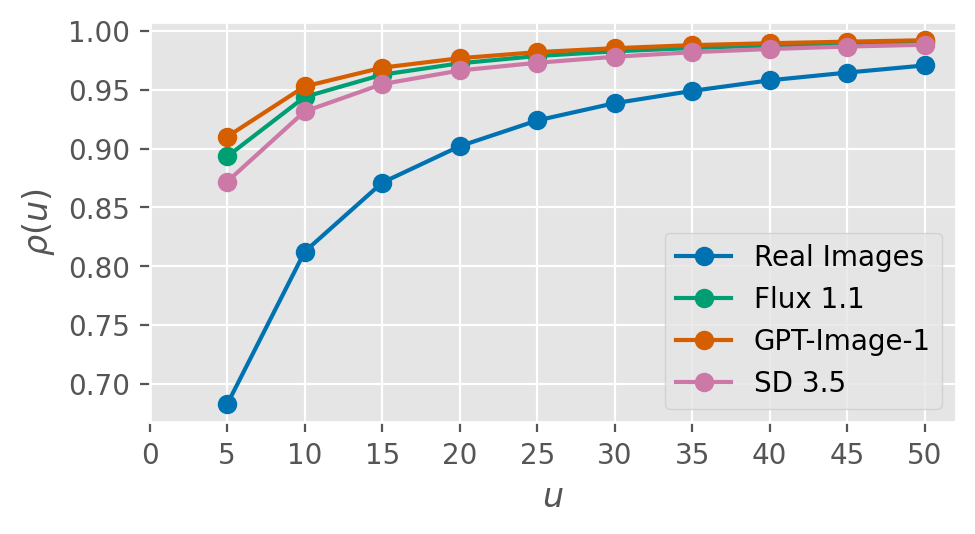}

   \caption{Concept representation intra-similarity scores $\rho(u)$, computed both for the real concept images and the synthetic ones using Visual-TCAV. Increasing the number of concept images $u$ results in greater intra-similarity, a consistent trend across both real and synthetic concepts. Synthetic ones exhibit higher intra-similarity even with a few concept images, suggesting that zero-shot T2I models in this setting may have limited capacity to capture real-world concept diversity.}
   \label{fig:intra_similarity}
\end{figure}

We conduct this experiment to compare the behavior of $\mathbf{v}_c^{\text{gen}}$ and $\mathbf{v}_c^{\text{real}}$ as their concept image subset sizes $u$ change, using the intra-similarity as a measure of convergence.
For synthetic concepts, a significant difference indicates misalignment, while excessively high similarity suggests that synthetic images fail to capture the concepts' complexity.

As illustrated in \Cref{fig:intra_similarity}, increasing the number of concept images reduces variability, a trend observed for both real and synthetic CAVs. However, even when the number of concept images is limited, synthetic CAVs are consistently higher than real ones ($\rho_c^{\text{gen}}(u) > \rho_c^{\text{real}}(u) \; \forall u$). Considering RQ2, this behavior highlights the limitations of the tested zero-shot T2I models in capturing the intrinsic diversity of concepts, underscoring their reduced ability to accurately reproduce real-world scenarios.

We also compute $\rho_c^{\text{CLIP,real}}$ and $\rho_c^{\text{CLIP,gen}}$ using CLIP, which are independent from $u$, whose results on average are 0.743 ± 0.093 for real images, and 0.858 ± 0.079 for Flux 1.1, 0.911 ± 0.079 for GPT-I-1, and 0.876 ± 0.076 for SD 3.5, confirming the pattern presented above.

\subsection{Impact on Downstream Explanation Task}

\begin{figure}[t]
  \centering
   \includegraphics[width=1\linewidth]{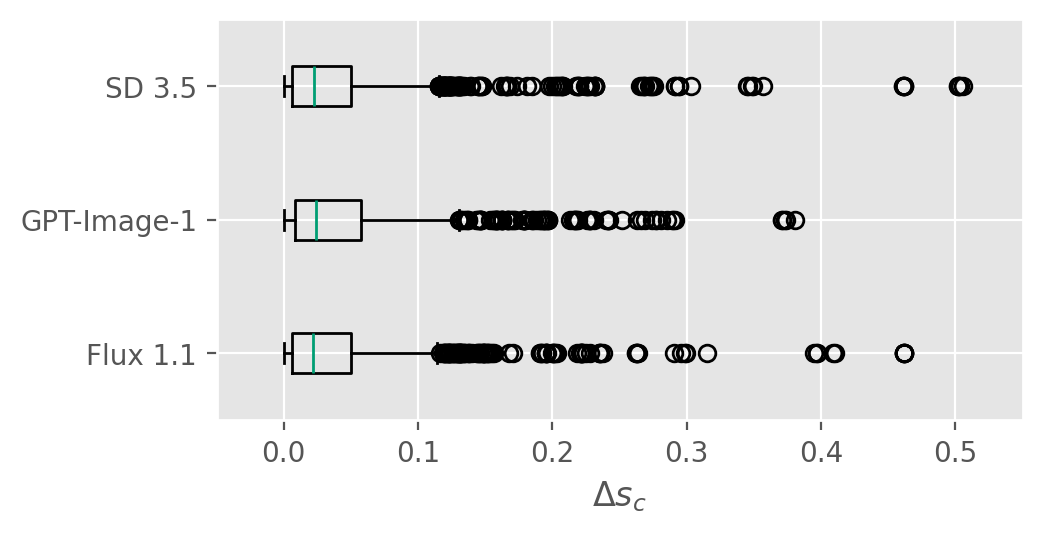}

   \caption{Boxplot showing the distributions of variations in importance scores ($\Delta s_c$) across different T2I generative models, computed using Visual-TCAV. It highlights low average differences and high variability between real and synthetic concepts ($0.039\pm0.055$ for Flux, $0.043\pm0.052$ for GPT-I-1, and $0.041\pm0.061$ for SD 3.5). The accompanying KS test confirms that these importance distributions are statistically distinct ($p<10^{-6}$), underscoring the semantic divergence introduced by synthetic generation.}
   \label{fig:attr_change}
\end{figure}

We analyze the importance differences to assess how closely the synthetic representations reproduce the explanatory behavior of their real counterparts. To this end, for each $c$, we extract the importances of all images with respect to their target class $k_c$ and compute the absolute value in the change in the importance score $\Delta s_c$. \Cref{fig:attr_change} presents the distributions of this metric for the three T2I models. Overall, the absolute importance variations have a low average with high variability ($0.039\pm0.055$ for Flux, $0.043\pm0.052$ for GPT-I-1, and $0.041\pm0.061$ for SD 3.5), indicating differences between real and synthetic concepts. To further validate this observation, we perform a two-sided Kolmogorov-Smirnov (KS) test to assess whether the distributions of real and synthetic importances differ. For all three T2I models, we observe a significant variation with high confidence (KS statistics of $0.196$, $0.316$, and $0.245$, all with $p < 10^{-6}$), indicating that the importances differ. More details are presented in \Cref{sec:app_importances_vtcav,sec:app_importances_tcav}.

Considering RQ3, even though synthetic data can approximate the visual characteristics of concept images, the observed deviations indicate that they fail to replicate the explanatory behavior of real images.

\subsection{Counterfactual Testing on Downstream Task}

This part evaluates both the impact of synthetic concepts on ablated class images (RQ4) and the effectiveness of the I2I removal process itself.

\begin{figure}[t]
  \centering
   \includegraphics[width=1\linewidth]{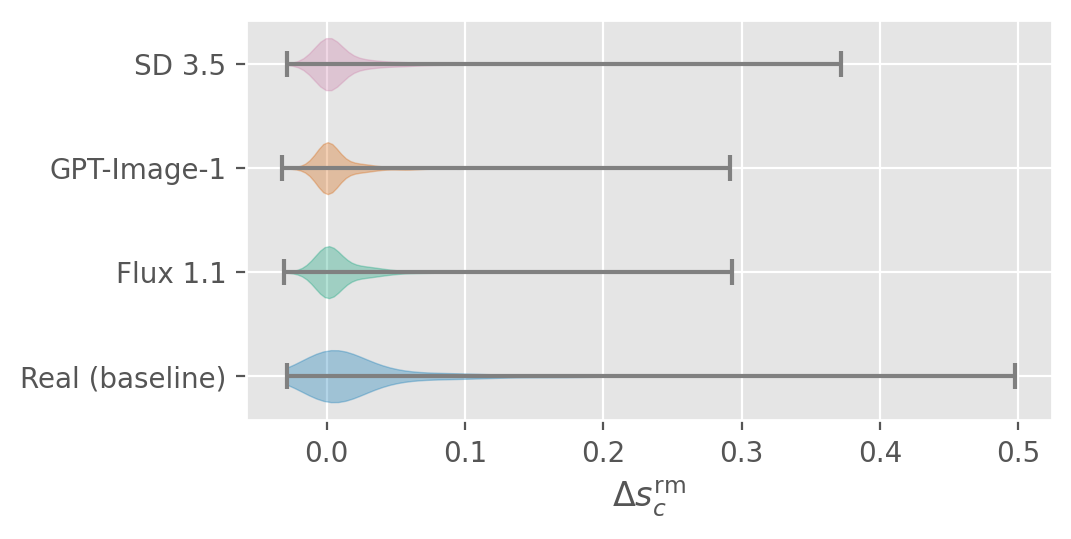}

   \caption{Violin plots illustrating the distribution of importance variations ($\Delta s_c^{\text{rm}}$) computed using Visual-TCAV after removing concepts from input class images using an I2I model. Most values cluster near zero for both real and generated concepts, but the latter have a narrower distribution (averages of $0.016\pm0.037$ for Flux, $0.013\pm0.032$ for GPT-I-1, and $0.016\pm0.04$ for SD 3.5) compared to the former ($0.031\pm0.065$). This highlights that concept ablation has a lower impact on synthetic importances.}
   \label{fig:kde_plot}
\end{figure}

\subsubsection{Concept Importance Variation After Removal}

Using the ablated class images, we compute the importance variations for both real ($\Delta s_c^{\text{real},\text{rm}}$) and synthetic concepts ($\Delta s_c^{\text{gen},\text{rm}}$). We compare these scores in \Cref{fig:kde_plot}, which presents the violin plots of their distributions. From the figure, we can observe a difference in the impact of the ablation procedure when using real or synthetic concepts. Specifically, removing a concept results in a lower change in the importance score when considering generated concept images (averages of $0.016\pm0.037$ for Flux, $0.013\pm0.032$ for GPT-I-1, and $0.016\pm0.04$ for SD 3.5) compared to the baseline of real concept images ($0.031\pm0.065$). Quantitatively, a two-sided KS test confirms that the distributions differ significantly ($p < 10^{-6}$).

\begin{figure}[t]
  \centering
   \includegraphics[width=1\linewidth]{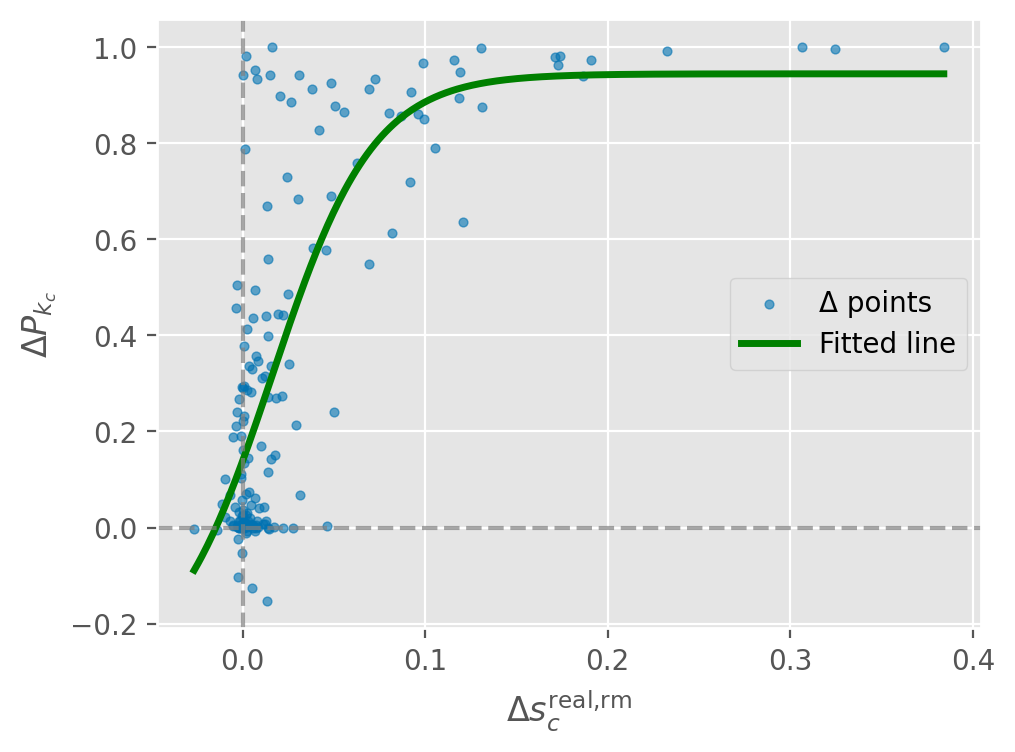}

   \caption{Scatter plot illustrating the relationship between the importance variation ($\Delta s_{c}^{\mathrm{real,rm}}$), computed with Visual-TCAV using real concept images on ablated class images, and the corresponding change in the model $f$’s classification probability ($\Delta P_{k_c}$). Spearman's rank correlation indicates a positive association ($\rho = 0.656$, $p < 10^{-6}$). A fitted logistic curve is overlaid ($RMSE = 0.240$), revealing an increasing monotonic trend.}
   \label{fig:attr_pred_ablation}
\end{figure}

\subsubsection{Concept Removal Validation via Accuracy Drop}

Finally, to evaluate the effectiveness of the concept removal procedure, we analyze whether a monotonic relationship exists between the importance variation $\Delta s_{c}^\text{real,rm}$ and the corresponding drop in classification probability $\Delta P_{k_c}$. To this end, we compute these quantities across all benchmarked concepts and image classifiers. In this scenario, we observe a monotonic trend where higher variations in importance scores are associated with higher drops in the predicted probability (Spearman Rank Correlation $\rho = 0.656$, $p < 10^{-6}$). This behavior is shown in \Cref{fig:attr_pred_ablation}.

\section{Conclusion}
\label{sec:conclusion}

This work presents a concept-generation framework that integrates zero-shot T2I generative models with predefined prompts into user-defined, concept-based XAI methods, aiming to assess whether automatically generated concepts can substitute for real ones while preserving their explanatory behavior. To this end, we evaluate (RQ1) the alignment between concept representations derived from synthetic and real images for the same semantic concept, (RQ2) how increasing the number of concept images affects concept representation intra-similarity across real and synthetic sets, (RQ3) the impact of replacing real concept images with synthetic ones on downstream explanation tasks, and (RQ4) how concept removal via prompt-driven image editing alters importance scores in counterfactual explanations for both real and generated images.

The results obtained provide evidence that, for the considered T2I models, synthetic concepts generated using a zero-shot approach may have a limited ability to reproduce key aspects of real concepts. Specifically, RQ1 confirmed that generators achieve systematic but partial alignment of latent representations, RQ2 demonstrated that synthetic concepts may struggle in capturing the intrinsic diversity of real-world concepts, RQ3 showed that a significant variation exists between synthetic attribution scores and real ones for the selected relevant classes, while RQ4 found that concept removal causes a smaller shift in attribution scores for generated images than for real ones.
This emphasizes the importance of carefully evaluating the use of synthetic concepts in XAI pipelines.

Future work will include coupling our T2I generation pipeline with fine-tuning techniques or I2I generative models, such as few-shot learning. Additionally, the discussed methodology could be applied to domain-specific concepts to evaluate generative models' capabilities on more specialized tasks.
{
    \small
    \bibliographystyle{ieeenat_fullname.bst}
    \bibliography{main}

@String(CVPR= {IEEE Conf. Comput. Vis. Pattern Recog.})

@String(ECCV= {Eur. Conf. Comput. Vis.})

@String(CVPRW= {IEEE Conf. Comput. Vis. Pattern Recog. Worksh.})

@String(CVPR  = {CVPR})

@String(ECCV  = {ECCV})

@String(CVPRW= {CVPRW})

@article{arrieta,
title = {Explainable Artificial Intelligence (XAI): Concepts, taxonomies, opportunities and challenges toward responsible AI},
journal = {Information Fusion},
volume = {58},
pages = {82-115},
year = {2020},
issn = {1566-2535},
doi = {https://doi.org/10.1016/j.inffus.2019.12.012},
url = {https://www.sciencedirect.com/science/article/pii/S1566253519308103},
author = {Alejandro {Barredo Arrieta} and Natalia Díaz-Rodríguez and Javier {Del Ser} and Adrien Bennetot and Siham Tabik and Alberto Barbado and Salvador Garcia and Sergio Gil-Lopez and Daniel Molina and Richard Benjamins and Raja Chatila and Francisco Herrera}
}

@incollection{XAI,
title = {2 - Foundational approaches to post-hoc explainability for image classification},
editor = {William Lawless and Ranjeev Mittu and Donald Sofge and Marco Brambilla},
booktitle = {Bi-directionality in Human-AI Collaborative Systems},
publisher = {Academic Press},
pages = {23-54},
year = {2025},
isbn = {978-0-443-40553-2},
doi = {https://doi.org/10.1016/B978-0-44-340553-2.00008-3},
url = {https://www.sciencedirect.com/science/article/pii/B9780443405532000083},
author = {Antonio {De Santis} and Riccardo Campi and Matteo Bianchi and Andrea Tocchetti and Marco Brambilla}
}

@InProceedings{text2concept,
    author    = {Moayeri, Mazda and Rezaei, Keivan and Sanjabi, Maziar and Feizi, Soheil},
    title     = {Text2Concept: Concept Activation Vectors Directly From Text},
    booktitle = {Proceedings of the IEEE/CVF Conference on Computer Vision and Pattern Recognition (CVPR) Workshops},
    month     = {June},
    year      = {2023},
    pages     = {3744-3749}
}

@InProceedings{clip,
  title = 	 {Learning Transferable Visual Models From Natural Language Supervision},
  author =       {Radford, Alec and Kim, Jong Wook and Hallacy, Chris and Ramesh, Aditya and Goh, Gabriel and Agarwal, Sandhini and Sastry, Girish and Askell, Amanda and Mishkin, Pamela and Clark, Jack and Krueger, Gretchen and Sutskever, Ilya},
  booktitle = 	 {Proceedings of the 38th International Conference on Machine Learning},
  pages = 	 {8748--8763},
  year = 	 {2021},
  editor = 	 {Meila, Marina and Zhang, Tong},
  volume = 	 {139},
  series = 	 {Proceedings of Machine Learning Research},
  month = 	 {18--24 Jul},
  publisher =    {PMLR},
  url = 	 {https://proceedings.mlr.press/v139/radford21a.html}
}

@misc{vtcav,
	title = {Visual-{TCAV}: {Concept}-based {Attribution} and {Saliency} {Maps} for {Post}-hoc {Explainability} in {Image} {Classification}},
	shorttitle = {Visual-{TCAV}},
	url = {http://arxiv.org/abs/2411.05698},
	doi = {10.48550/arXiv.2411.05698},
	language = {en},
	urldate = {2025-10-01},
	publisher = {arXiv},
	author = {Santis, Antonio De and Campi, Riccardo and Bianchi, Matteo and Brambilla, Marco},
	month = jun,
	year = {2025},
	note = {arXiv:2411.05698 [cs]},
}

@InProceedings{tcav,
  title = 	 {Interpretability Beyond Feature Attribution: Quantitative Testing with Concept Activation Vectors ({TCAV})},
  author =       {Kim, Been and Wattenberg, Martin and Gilmer, Justin and Cai, Carrie and Wexler, James and Viegas, Fernanda and sayres, Rory},
  booktitle = 	 {Proceedings of the 35th International Conference on Machine Learning},
  pages = 	 {2668--2677},
  year = 	 {2018},
  editor = 	 {Dy, Jennifer and Krause, Andreas},
  volume = 	 {80},
  series = 	 {Proceedings of Machine Learning Research},
  month = 	 {10--15 Jul},
  publisher =    {PMLR},
  url = 	 {https://proceedings.mlr.press/v80/kim18d.html}
}

@inproceedings{patterncav,
title={Navigating Neural Space: Revisiting Concept Activation Vectors to Overcome Directional Divergence},
author={Frederik Pahde and Maximilian Dreyer and Moritz Weckbecker and Leander Weber and Christopher J. Anders and Thomas Wiegand and Wojciech Samek and Sebastian Lapuschkin},
booktitle={The Thirteenth International Conference on Learning Representations},
year={2025},
url={https://openreview.net/forum?id=Q95MaWfF4e}
}

@phdthesis{MartinPhd,
  author = {Tyler Martin and Adrian Weller},
  title = "{Interpretable Machine Learning}",
  school = "Dept.\ of Engineering, University of Cambridge",
  year = 2019,
  type = "{M.Phil.} diss.",
  month = "August",
  url = {https://www.mlmi.eng.cam.ac.uk/files/tam_final_reduced.pdf}
}

@book{molnar2022,
  title     = "Interpretable Machine Learning: A Guide For Making Black Box Models Explainable",
  author    = "Molnar, Christoph",
  year      = 2022,
  publisher = "Independently published"
}

@INPROCEEDINGS{SCG,
  author={Campi, Riccardo and Borrego, Santiago and De Santis, Antonio and Bianchi, Matteo and Tocchetti, Andrea and Brambilla, Marco},
  booktitle={2025 IEEE/CVF Conference on Computer Vision and Pattern Recognition Workshops (CVPRW)}, 
  title={Towards Synthetic Concept Activation Vectors via Generative Models}, 
  year={2025},
  volume={},
  number={},
  pages={2711-2719},
  doi={10.1109/CVPRW67362.2025.00256}}

@inproceedings{LGCAV,
 author = {Huang, Qihan and Song, Jie and Xue, Mengqi and Zhang, Haofei and Hu, Bingde and Wang, Huiqiong and Jiang, Hao and Wang, Xingen and Song, Mingli},
 booktitle = {Advances in Neural Information Processing Systems},
 doi = {10.52202/079017-1249},
 editor = {A. Globerson and L. Mackey and D. Belgrave and A. Fan and U. Paquet and J. Tomczak and C. Zhang},
 pages = {39522--39551},
 publisher = {Curran Associates, Inc.},
 title = {LG-CAV: Train Any Concept Activation Vector with Language Guidance},
 url = {https://proceedings.neurips.cc/paper_files/paper/2024/file/45d4924460c37853d57885d8af0b8d5c-Paper-Conference.pdf},
 volume = {37},
 year = {2024}
}

@InProceedings{DTD,
	      Author    = {M. Cimpoi and S. Maji and I. Kokkinos and S. Mohamed and and A. Vedaldi},
	      Title     = {Describing Textures in the Wild},
	      Booktitle = {Proceedings of the {IEEE} Conf. on Computer Vision and Pattern Recognition ({CVPR})},
	      Year      = {2014}
}

@INPROCEEDINGS{imagenet,
  author={Deng, Jia and Dong, Wei and Socher, Richard and Li, Li-Jia and Kai Li and Li Fei-Fei},
  booktitle={2009 IEEE Conference on Computer Vision and Pattern Recognition}, 
  title={ImageNet: A large-scale hierarchical image database}, 
  year={2009},
  volume={},
  number={},
  pages={248-255},
  doi={10.1109/CVPR.2009.5206848}
}

@article{FMD,
   author = {Lavanya Sharan and Ruth Rosenholtz and Edward H. Adelson},
   title = {Accuracy and speed of material categorization in real-world images},
   journal = {Journal of Vision},
   volume = {14},
   number = {10},
   year = {2014}
  }

@InProceedings{resnetv2,
author="He, Kaiming
and Zhang, Xiangyu
and Ren, Shaoqing
and Sun, Jian",
editor="Leibe, Bastian
and Matas, Jiri
and Sebe, Nicu
and Welling, Max",
title="Identity Mappings in Deep Residual Networks",
booktitle="Computer Vision -- ECCV 2016",
year="2016",
publisher="Springer International Publishing",
address="Cham",
pages="630--645",
isbn="978-3-319-46493-0"
}

@INPROCEEDINGS{inceptionv3,
    author={Szegedy, Christian and Vanhoucke, Vincent and Ioffe, Sergey and Shlens, Jon and Wojna, Zbigniew},
    booktitle={2016 IEEE Conference on Computer Vision and Pattern Recognition (CVPR)}, 
    title={Rethinking the Inception Architecture for Computer Vision}, 
    year={2016},
    volume={},
    number={},
    pages={2818-2826},
    doi={10.1109/CVPR.2016.308}
}

@InProceedings{vgg,
  author       = "Karen Simonyan and Andrew Zisserman",
  title        = "Very Deep Convolutional Networks for Large-Scale Image Recognition",
  booktitle    = "International Conference on Learning Representations",
  year         = "2015",
}

@INPROCEEDINGS{convnext,
  author={Liu, Zhuang and Mao, Hanzi and Wu, Chao-Yuan and Feichtenhofer, Christoph and Darrell, Trevor and Xie, Saining},
  booktitle={2022 IEEE/CVF Conference on Computer Vision and Pattern Recognition (CVPR)}, 
  title={A ConvNet for the 2020s}, 
  year={2022},
  volume={},
  number={},
  pages={11966-11976},
  doi={10.1109/CVPR52688.2022.01167}}

@misc{flux,
      title={FLUX.1 Kontext: Flow Matching for In-Context Image Generation and Editing in Latent Space},
      author={Black Forest Labs and Stephen Batifol and Andreas Blattmann and Frederic Boesel and Saksham Consul and Cyril Diagne and Tim Dockhorn and Jack English and Zion English and Patrick Esser and Sumith Kulal and Kyle Lacey and Yam Levi and Cheng Li and Dominik Lorenz and Jonas Müller and Dustin Podell and Robin Rombach and Harry Saini and Axel Sauer and Luke Smith},
      year={2025},
      eprint={2506.15742},
      archivePrefix={arXiv},
      primaryClass={cs.GR},
      url={https://arxiv.org/abs/2506.15742},
}

@inproceedings{sd3,
author = {Esser, Patrick and Kulal, Sumith and Blattmann, Andreas and Entezari, Rahim and M\"{u}ller, Jonas and Saini, Harry and Levi, Yam and Lorenz, Dominik and Sauer, Axel and Boesel, Frederic and Podell, Dustin and Dockhorn, Tim and English, Zion and Rombach, Robin},
title = {Scaling rectified flow transformers for high-resolution image synthesis},
year = {2024},
publisher = {JMLR.org},
booktitle = {Proceedings of the 41st International Conference on Machine Learning},
articleno = {503},
numpages = {28},
location = {Vienna, Austria},
series = {ICML'24}
}

@INPROCEEDINGS{perceptual_similarity,
  author={Zhang, Richard and Isola, Phillip and Efros, Alexei A. and Shechtman, Eli and Wang, Oliver},
  booktitle={2018 IEEE/CVF Conference on Computer Vision and Pattern Recognition}, 
  title={The Unreasonable Effectiveness of Deep Features as a Perceptual Metric}, 
  year={2018},
  volume={},
  number={},
  pages={586-595},
  doi={10.1109/CVPR.2018.00068}}
}

\clearpage
\setcounter{page}{1}
\maketitlesupplementary

\setcounter{section}{0}
\renewcommand{\thesection}{\Alph{section}}

\vspace*{\fill}

\section{Complete List of Concepts}
\label{sec:app_concepts}

Table~\ref{tab:concept_summary} contains a detailed overview of the concept-class dataset used in our study.
The dataset comprises 41 concepts, each mapped to a unique class label derived from the ImageNet dataset. Real concept images are drawn from four distinct datasets: DTD, ImageNet, FMD, and search engine results.

\begin{table}[!h]
    \centering
    \begin{tabular}{cccc|cccc}
        \toprule
        \textbf{Type} & \textbf{Dataset} & \textbf{Concept} & \textbf{Relevant Class} & \textbf{Type} & \textbf{Dataset} & \textbf{Concept} & \textbf{Relevant Class} \\
        \midrule
        Texture & DTD & Bubbly & Beer glass &           Object & ImageNet & Asparagus & Stone wall \\
        Texture & DTD & Chequered & Crossword puzzle &  Object & ImageNet & Cloister & Church \\
        Texture & DTD & Cracked & Stone wall &          Object & ImageNet & Feline & Leopard \\
        Texture & DTD & Crystalline & Honeycomb &       Object & ImageNet & Grandfather & Tench \\
        Texture & DTD & Dotted & Dalmatian &            Object & ImageNet & Guitarist & Electric guitar \\
        Texture & DTD & Frilly & Overskirt &            Object & ImageNet & Kitchen & Waffle iron \\
        Texture & DTD & Honeycombed & Honeycomb &       Object & ImageNet & Lichen & Stone wall \\
        Texture & DTD & Meshed & Honeycomb &            Object & ImageNet & Rodent & Leopard \\
        Texture & DTD & Perforated & Honeycomb &        Object & ImageNet & Steeple & Church \\
        Texture & DTD & Striped & Zebra &               Object & ImageNet & Tattoo & Zebra \\
        Texture & DTD & Waffled & Waffle iron &         Object & Search & Cast iron & Waffle iron \\
        Texture & DTD & Woven & Overskirt &             Object & Search & Fin & Tench \\
        Texture & DTD & Wrinkled & Overskirt &          Object & Search & Grass & Golf ball \\
        Texture & FMD & Fabric & Trench coat &          Object & Search & Newspaper & Crossword puzzle \\
        Texture & FMD & Foliage & Spider web &          Object & Search & Pen & Fountain pen \\
        Texture & FMD & Glass & Beer glass &            Object & Search & Sphere & Golf ball \\
        Texture & FMD & Leather & Trench coat &         Texture & Search & Leopard print & Leopard \\
        Texture & FMD & Metal & Waffle iron &           Texture & Search & Spotted & Dalmatian \\
        Texture & FMD & Paper & Crossword puzzle &       & & & \\
        Texture & FMD & Plastic & Golf ball &            & & & \\
        Texture & FMD & Stone & Stone wall               & & & \\
        Texture & FMD & Water & Beer glass               & & & \\
        Texture & FMD & Wood & Wooden spoon               & & & \\
        \bottomrule
    \end{tabular}
    \caption{Summary of the concept-class dataset used in our experiments. Each concept is associated with a specific class (e.g., dotted $\rightarrow$ dalmatian). Concepts are drawn from four distinct datasets: Describable Textures Dataset (DTD), ImageNet, Flickr Material Database (FMD), and search engine results.}
    \label{tab:concept_summary}
\end{table}

\vspace*{\fill}

\newpage

\section{Average Cosine Similarity per Concept using Visual-TCAV}
\label{sec:app_cos_sim_vtcav}

\vspace*{\fill}

\begin{table}[h!]
\footnotesize
\centering
\begin{tabular}{|c|c|c|c|c|c|c|}
\toprule
Class & \multicolumn{3}{c|}{beer\_glass} & \multicolumn{2}{c|}{church} & crossword\_puzzle \\
\midrule
Concept & bubbly & glass & water & cloister & steeple & chequered \\
\midrule
Flux & $0.666\pm 0.14$ & $0.453\pm 0.139$ & $0.59\pm 0.069$ & $0.848\pm 0.017$ & $0.726\pm 0.07$ & $0.217\pm 0.141$ \\
GPT-Image-1 & $0.659\pm 0.038$ & $0.389\pm 0.162$ & $0.509\pm 0.14$ & $0.389\pm 0.081$ & $0.842\pm 0.042$ & $0.586\pm 0.096$ \\
SD 3.5 & $0.689\pm 0.1$ & $0.395\pm 0.103$ & $0.523\pm 0.099$ & $0.894\pm 0.026$ & $0.764\pm 0.057$ & $0.664\pm 0.102$ \\
\bottomrule
\end{tabular}

\begin{tabular}{|c|c|c|c|c|c|c|}
\toprule
Class & \multicolumn{2}{c|}{crossword\_puzzle} & \multicolumn{2}{c|}{dalmatian} & electric\_guitar & fountain\_pen \\
\midrule
Concept & newspaper & paper & dotted & spotted & guitarist & pen \\
\midrule
Flux & $0.589\pm 0.052$ & $0.589\pm 0.067$ & $0.376\pm 0.126$ & $0.375\pm 0.155$ & $0.815\pm 0.04$ & $0.595\pm 0.206$ \\
GPT-Image-1 & $0.702\pm 0.042$ & $0.475\pm 0.061$ & $0.654\pm 0.155$ & $0.554\pm 0.144$ & $0.699\pm 0.065$ & $0.305\pm 0.174$ \\
SD 3.5 & $0.69\pm 0.125$ & $0.533\pm 0.074$ & $0.525\pm 0.098$ & $0.346\pm 0.138$ & $0.818\pm 0.036$ & $0.482\pm 0.226$ \\
\bottomrule
\end{tabular}

\begin{tabular}{|c|c|c|c|c|c|c|}
\toprule
Class & \multicolumn{3}{c|}{golf\_ball} & \multicolumn{3}{c|}{honeycomb} \\
\midrule
Concept & grass & plastic & sphere & crystalline & honeycombed & meshed \\
\midrule
Flux & $0.411\pm 0.168$ & $0.504\pm 0.092$ & $0.483\pm 0.111$ & $0.486\pm 0.083$ & $0.707\pm 0.188$ & $0.588\pm 0.123$ \\
GPT-Image-1 & $0.328\pm 0.194$ & $0.247\pm 0.085$ & $0.487\pm 0.094$ & $0.482\pm 0.094$ & $0.787\pm 0.07$ & $0.411\pm 0.057$ \\
SD 3.5 & $0.485\pm 0.131$ & $0.435\pm 0.096$ & $0.561\pm 0.103$ & $0.498\pm 0.153$ & $0.866\pm 0.045$ & $0.571\pm 0.145$ \\
\bottomrule
\end{tabular}

\begin{tabular}{|c|c|c|c|c|c|c|}
\toprule
Class & honeycomb & \multicolumn{3}{c|}{leopard} & \multicolumn{2}{c|}{overskirt} \\
\midrule
Concept & perforated & feline & leopard\_print & rodent & frilly & woven \\
\midrule
Flux & $0.336\pm 0.123$ & $0.537\pm 0.132$ & $0.6\pm 0.127$ & $0.607\pm 0.072$ & $0.473\pm 0.142$ & $0.456\pm 0.182$ \\
GPT-Image-1 & $0.658\pm 0.064$ & $0.732\pm 0.062$ & $0.828\pm 0.13$ & $0.685\pm 0.117$ & $0.512\pm 0.128$ & $0.378\pm 0.082$ \\
SD 3.5 & $0.641\pm 0.045$ & $0.612\pm 0.114$ & $0.682\pm 0.1$ & $0.726\pm 0.095$ & $0.432\pm 0.151$ & $0.691\pm 0.066$ \\
\bottomrule
\end{tabular}

\begin{tabular}{|c|c|c|c|c|c|c|}
\toprule
Class & overskirt & spider\_web & \multicolumn{4}{c|}{stone\_wall} \\
\midrule
Concept & wrinkled & foliage & asparagus & cracked & lichen & stone \\
\midrule
Flux & $0.535\pm 0.133$ & $0.568\pm 0.192$ & $0.622\pm 0.124$ & $0.657\pm 0.041$ & $0.382\pm 0.312$ & $0.542\pm 0.099$ \\
GPT-Image-1 & $0.419\pm 0.106$ & $0.628\pm 0.17$ & $0.279\pm 0.223$ & $0.696\pm 0.057$ & $0.505\pm 0.115$ & $0.559\pm 0.176$ \\
SD 3.5 & $0.559\pm 0.14$ & $0.508\pm 0.217$ & $0.693\pm 0.078$ & $0.678\pm 0.116$ & $0.569\pm 0.027$ & $0.517\pm 0.144$ \\
\bottomrule
\end{tabular}

\begin{tabular}{|c|c|c|c|c|c|c|}
\toprule
Class & \multicolumn{2}{c|}{tench} & \multicolumn{2}{c|}{trench\_coat} & \multicolumn{2}{c|}{waffle\_iron} \\
\midrule
Concept & fin & grandfather & fabric & leather & cast\_iron & kitchen \\
\midrule
Flux & $0.151\pm 0.084$ & $0.666\pm 0.085$ & $0.572\pm 0.111$ & $0.602\pm 0.124$ & $0.164\pm 0.114$ & $0.773\pm 0.117$ \\
GPT-Image-1 & $0.282\pm 0.097$ & $0.658\pm 0.155$ & $0.508\pm 0.114$ & $0.572\pm 0.129$ & $0.31\pm 0.104$ & $0.445\pm 0.086$ \\
SD 3.5 & $0.317\pm 0.118$ & $0.681\pm 0.099$ & $0.55\pm 0.102$ & $0.567\pm 0.111$ & $0.282\pm 0.144$ & $0.715\pm 0.069$ \\
\bottomrule
\end{tabular}

\begin{tabular}{|c|c|c|c|c|c|}
\toprule
Class & \multicolumn{2}{c|}{waffle\_iron} & wooden\_spoon & \multicolumn{2}{c|}{zebra} \\
\midrule
Concept & metal & waffled & wood & striped & tattoo \\
\midrule
Flux & $0.458\pm 0.134$ & $0.81\pm 0.047$ & $0.542\pm 0.129$ & $0.583\pm 0.236$ & $0.525\pm 0.058$ \\
GPT-Image-1 & $0.502\pm 0.108$ & $0.826\pm 0.029$ & $0.656\pm 0.131$ & $0.624\pm 0.149$ & $0.345\pm 0.088$ \\
SD 3.5 & $0.399\pm 0.206$ & $0.869\pm 0.037$ & $0.572\pm 0.096$ & $0.335\pm 0.151$ & $0.755\pm 0.045$ \\
\bottomrule
\end{tabular}

\label{tab:intra_sim_all}
\caption{Average cosine similarities between real and generated concepts whose CAVs are computed using Visual-TCAV. Results are aggregated by averaging across four architectures and two layers each (i.e., the outputs of the final two blocks for ResNet-50-V2, Inception-V3, and ConvNeXt, and the outputs of the last two convolutional layers of VGG-16).}
\end{table}

\vspace*{\fill}\newpage

\section{Average Cosine Similarity per Concept using TCAV}
\label{sec:app_cos_sim_tcav}

\vspace*{\fill}

\begin{table}[h!]
\footnotesize
\centering

\begin{tabular}{|c|c|c|c|c|c|c|}
\toprule
Class & \multicolumn{3}{c|}{beer\_glass} & \multicolumn{2}{c|}{church} & crossword\_puzzle \\
\midrule
Concept & bubbly & glass & water & cloister & steeple & chequered \\
\midrule
Flux & $0.416\pm 0.137$ & $0.275\pm 0.06$ & $0.438\pm 0.049$ & $0.689\pm 0.065$ & $0.515\pm 0.108$ & $0.151\pm 0.098$ \\
GPT-Image-1 & $0.563\pm 0.056$ & $0.249\pm 0.046$ & $0.517\pm 0.082$ & $0.11\pm 0.032$ & $0.779\pm 0.047$ & $0.542\pm 0.051$ \\
SD 3.5 & $0.526\pm 0.1$ & $0.258\pm 0.086$ & $0.402\pm 0.056$ & $0.74\pm 0.078$ & $0.742\pm 0.056$ & $0.746\pm 0.025$ \\
\bottomrule
\end{tabular}

\centering
\begin{tabular}{|c|c|c|c|c|c|c|}
\toprule
Class & \multicolumn{2}{c|}{crossword\_puzzle} & \multicolumn{2}{c|}{dalmatian} & electric\_guitar & fountain\_pen \\
\midrule
Concept & newspaper & paper & dotted & spotted & guitarist & pen \\
\midrule
Flux & $0.535\pm 0.055$ & $0.4\pm 0.068$ & $0.448\pm 0.06$ & $0.259\pm 0.119$ & $0.693\pm 0.098$ & $0.655\pm 0.11$ \\
GPT-Image-1 & $0.541\pm 0.04$ & $0.376\pm 0.056$ & $0.597\pm 0.099$ & $0.443\pm 0.12$ & $0.564\pm 0.072$ & $0.237\pm 0.171$ \\
SD 3.5 & $0.759\pm 0.041$ & $0.426\pm 0.037$ & $0.568\pm 0.069$ & $0.395\pm 0.133$ & $0.724\pm 0.094$ & $0.539\pm 0.072$ \\
\bottomrule
\end{tabular}

\centering
\begin{tabular}{|c|c|c|c|c|c|c|}
\toprule
Class & \multicolumn{3}{c|}{golf\_ball} & \multicolumn{3}{c|}{honeycomb} \\
\midrule
Concept & grass & plastic & sphere & crystalline & honeycombed & meshed \\
\midrule
Flux & $0.401\pm 0.159$ & $0.428\pm 0.049$ & $0.282\pm 0.036$ & $0.413\pm 0.052$ & $0.802\pm 0.05$ & $0.584\pm 0.061$ \\
GPT-Image-1 & $0.285\pm 0.181$ & $0.166\pm 0.134$ & $0.263\pm 0.071$ & $0.399\pm 0.05$ & $0.77\pm 0.096$ & $0.271\pm 0.071$ \\
SD 3.5 & $0.444\pm 0.131$ & $0.362\pm 0.051$ & $0.413\pm 0.049$ & $0.376\pm 0.043$ & $0.853\pm 0.037$ & $0.659\pm 0.063$ \\
\bottomrule
\end{tabular}

\centering
\begin{tabular}{|c|c|c|c|c|c|c|}
\toprule
Class & honeycomb & \multicolumn{3}{c|}{leopard} & \multicolumn{2}{c|}{overskirt} \\
\midrule
Concept & perforated & feline & leopard\_print & rodent & frilly & woven \\
\midrule
Flux & $0.372\pm 0.053$ & $0.387\pm 0.104$ & $0.383\pm 0.256$ & $0.296\pm 0.231$ & $0.431\pm 0.09$ & $0.405\pm 0.14$ \\
GPT-Image-1 & $0.61\pm 0.071$ & $0.644\pm 0.053$ & $0.878\pm 0.016$ & $0.477\pm 0.086$ & $0.454\pm 0.106$ & $0.236\pm 0.102$ \\
SD 3.5 & $0.572\pm 0.065$ & $0.485\pm 0.047$ & $0.57\pm 0.202$ & $0.575\pm 0.107$ & $0.413\pm 0.102$ & $0.706\pm 0.038$ \\
\bottomrule
\end{tabular}

\centering
\begin{tabular}{|c|c|c|c|c|c|c|}
\toprule
Class & overskirt & spider\_web & \multicolumn{4}{c|}{stone\_wall} \\
\midrule
Concept & wrinkled & foliage & asparagus & cracked & lichen & stone \\
\midrule
Flux & $0.31\pm 0.061$ & $0.366\pm 0.12$ & $0.545\pm 0.071$ & $0.585\pm 0.071$ & $0.255\pm 0.133$ & $0.309\pm 0.057$ \\
GPT-Image-1 & $0.313\pm 0.058$ & $0.676\pm 0.048$ & $0.107\pm 0.105$ & $0.496\pm 0.257$ & $0.523\pm 0.038$ & $0.549\pm 0.045$ \\
SD 3.5 & $0.488\pm 0.06$ & $0.368\pm 0.101$ & $0.631\pm 0.093$ & $0.666\pm 0.073$ & $0.666\pm 0.055$ & $0.318\pm 0.125$ \\
\bottomrule
\end{tabular}

\centering
\begin{tabular}{|c|c|c|c|c|c|c|}
\toprule
Class & \multicolumn{2}{c|}{tench} & \multicolumn{2}{c|}{trench\_coat} & \multicolumn{2}{c|}{waffle\_iron} \\
\midrule
Concept & fin & grandfather & fabric & leather & cast\_iron & kitchen \\
\midrule
Flux & $0.08\pm 0.046$ & $0.595\pm 0.059$ & $0.565\pm 0.081$ & $0.542\pm 0.11$ & $0.162\pm 0.065$ & $0.797\pm 0.016$ \\
GPT-Image-1 & $0.116\pm 0.071$ & $0.541\pm 0.095$ & $0.523\pm 0.051$ & $0.528\pm 0.041$ & $0.161\pm 0.078$ & $0.25\pm 0.053$ \\
SD 3.5 & $0.099\pm 0.037$ & $0.515\pm 0.062$ & $0.562\pm 0.041$ & $0.537\pm 0.095$ & $0.245\pm 0.166$ & $0.653\pm 0.03$ \\
\bottomrule
\end{tabular}

\centering
\begin{tabular}{|c|c|c|c|c|c|}
\toprule
Class & \multicolumn{2}{c|}{waffle\_iron} & wooden\_spoon & \multicolumn{2}{c|}{zebra} \\
\midrule
Concept & metal & waffled & wood & striped & tattoo \\
\midrule
Flux & $0.282\pm 0.022$ & $0.759\pm 0.059$ & $0.333\pm 0.103$ & $0.462\pm 0.092$ & $0.444\pm 0.033$ \\
GPT-Image-1 & $0.411\pm 0.042$ & $0.818\pm 0.075$ & $0.582\pm 0.053$ & $0.628\pm 0.089$ & $0.193\pm 0.088$ \\
SD 3.5 & $0.398\pm 0.036$ & $0.835\pm 0.043$ & $0.395\pm 0.075$ & $0.376\pm 0.132$ & $0.662\pm 0.031$ \\
\bottomrule
\end{tabular}

\label{tab:cosine_sim_tcav_all}
\caption{Average cosine similarities between real and generated concepts whose CAVs are computed using TCAV. Results are aggregated by averaging across four architectures and two layers each (i.e., the outputs of the final two blocks for ResNet-50-V2, Inception-V3, and ConvNeXt, and the outputs of the last two convolutional layers of VGG-16).}
\end{table}

\vspace*{\fill}
\newpage

\section{Average Importances per Concept using VisualTCAV}
\label{sec:app_importances_vtcav}

\vspace*{\fill}

\begin{table}[h!]
\footnotesize
\centering
\begin{tabular}{|c|c|c|c|c|c|c|}
\toprule
Class & \multicolumn{3}{c|}{beer\_glass} & \multicolumn{2}{c|}{church} & crossword\_puzzle \\
\midrule
Concept & bubbly & glass & water & cloister & steeple & chequered \\
\midrule
Real & $0.027\pm 0.018$ & $0.104\pm 0.04$ & $0.042\pm 0.039$ & $0.07\pm 0.052$ & $0.111\pm 0.056$ & $0.086\pm 0.077$ \\
Flux & $0.011\pm 0.011$ & $0.081\pm 0.058$ & $0.03\pm 0.02$ & $0.075\pm 0.055$ & $0.112\pm 0.052$ & $0.02\pm 0.021$ \\
GPT-Image-1 & $0.006\pm 0.007$ & $0.085\pm 0.038$ & $0.001\pm 0.001$ & $0.015\pm 0.012$ & $0.082\pm 0.041$ & $0.022\pm 0.015$ \\
SD 3.5 & $0.012\pm 0.01$ & $0.069\pm 0.058$ & $0.003\pm 0.003$ & $0.054\pm 0.036$ & $0.067\pm 0.034$ & $0.076\pm 0.092$ \\
\bottomrule
\end{tabular}

\begin{tabular}{|c|c|c|c|c|c|c|}
\toprule
Class & \multicolumn{2}{c|}{crossword\_puzzle} & \multicolumn{2}{c|}{dalmatian} & electric\_guitar & fountain\_pen \\
\midrule
Concept & newspaper & paper & dotted & spotted & guitarist & pen \\
\midrule
Real & $0.123\pm 0.082$ & $0.039\pm 0.032$ & $0.012\pm 0.014$ & $0.171\pm 0.146$ & $0.149\pm 0.113$ & $0.04\pm 0.014$ \\
Flux & $0.049\pm 0.052$ & $0.058\pm 0.045$ & $0.004\pm 0.009$ & $0.032\pm 0.049$ & $0.134\pm 0.085$ & $0.057\pm 0.027$ \\
GPT-Image-1 & $0.093\pm 0.069$ & $0.013\pm 0.013$ & $0.005\pm 0.006$ & $0.027\pm 0.033$ & $0.089\pm 0.09$ & $0.032\pm 0.011$ \\
SD 3.5 & $0.089\pm 0.089$ & $0.035\pm 0.046$ & $0.0\pm 0.0$ & $0.018\pm 0.028$ & $0.167\pm 0.079$ & $0.055\pm 0.043$ \\
\bottomrule
\end{tabular}

\begin{tabular}{|c|c|c|c|c|c|c|}
\toprule
Class & \multicolumn{3}{c|}{golf\_ball} & \multicolumn{3}{c|}{honeycomb} \\
\midrule
Concept & grass & plastic & sphere & crystalline & honeycombed & meshed \\
\midrule
Real & $0.008\pm 0.01$ & $0.049\pm 0.028$ & $0.148\pm 0.107$ & $0.03\pm 0.026$ & $0.193\pm 0.136$ & $0.13\pm 0.124$ \\
Flux & $0.004\pm 0.008$ & $0.014\pm 0.013$ & $0.064\pm 0.038$ & $0.058\pm 0.04$ & $0.156\pm 0.073$ & $0.106\pm 0.117$ \\
GPT-Image-1 & $0.014\pm 0.011$ & $0.008\pm 0.01$ & $0.065\pm 0.025$ & $0.018\pm 0.019$ & $0.11\pm 0.074$ & $0.025\pm 0.035$ \\
SD 3.5 & $0.003\pm 0.002$ & $0.013\pm 0.016$ & $0.051\pm 0.031$ & $0.012\pm 0.006$ & $0.193\pm 0.121$ & $0.028\pm 0.019$ \\
\bottomrule
\end{tabular}

\begin{tabular}{|c|c|c|c|c|c|c|}
\toprule
Class & honeycomb & \multicolumn{3}{c|}{leopard} & \multicolumn{2}{c|}{overskirt} \\
\midrule
Concept & perforated & feline & leopard\_print & rodent & frilly & woven \\
\midrule
Real & $0.095\pm 0.07$ & $0.125\pm 0.074$ & $0.128\pm 0.063$ & $0.027\pm 0.029$ & $0.025\pm 0.022$ & $0.003\pm 0.004$ \\
Flux & $0.049\pm 0.034$ & $0.049\pm 0.04$ & $0.063\pm 0.083$ & $0.012\pm 0.011$ & $0.034\pm 0.031$ & $0.008\pm 0.011$ \\
GPT-Image-1 & $0.064\pm 0.05$ & $0.056\pm 0.053$ & $0.067\pm 0.056$ & $0.012\pm 0.016$ & $0.006\pm 0.009$ & $0.003\pm 0.003$ \\
SD 3.5 & $0.04\pm 0.024$ & $0.038\pm 0.032$ & $0.088\pm 0.076$ & $0.012\pm 0.015$ & $0.057\pm 0.035$ & $0.005\pm 0.008$ \\
\bottomrule
\end{tabular}

\begin{tabular}{|c|c|c|c|c|c|c|}
\toprule
Class & overskirt & spider\_web & \multicolumn{4}{c|}{stone\_wall} \\
\midrule
Concept & wrinkled & foliage & asparagus & cracked & lichen & stone \\
\midrule
Real & $0.02\pm 0.024$ & $0.015\pm 0.014$ & $0.02\pm 0.019$ & $0.074\pm 0.061$ & $0.058\pm 0.071$ & $0.1\pm 0.08$ \\
Flux & $0.03\pm 0.03$ & $0.005\pm 0.006$ & $0.004\pm 0.004$ & $0.049\pm 0.032$ & $0.006\pm 0.005$ & $0.067\pm 0.043$ \\
GPT-Image-1 & $0.003\pm 0.005$ & $0.003\pm 0.006$ & $0.002\pm 0.001$ & $0.034\pm 0.033$ & $0.006\pm 0.004$ & $0.027\pm 0.02$ \\
SD 3.5 & $0.013\pm 0.011$ & $0.008\pm 0.009$ & $0.003\pm 0.003$ & $0.03\pm 0.025$ & $0.013\pm 0.01$ & $0.063\pm 0.019$ \\
\bottomrule
\end{tabular}

\begin{tabular}{|c|c|c|c|c|c|c|}
\toprule
Class & \multicolumn{2}{c|}{tench} & \multicolumn{2}{c|}{trench\_coat} & \multicolumn{2}{c|}{waffle\_iron} \\
\midrule
Concept & fin & grandfather & fabric & leather & cast\_iron & kitchen \\
\midrule
Real & $0.055\pm 0.034$ & $0.036\pm 0.036$ & $0.028\pm 0.023$ & $0.015\pm 0.012$ & $0.13\pm 0.081$ & $0.019\pm 0.013$ \\
Flux & $0.002\pm 0.002$ & $0.013\pm 0.01$ & $0.018\pm 0.016$ & $0.007\pm 0.005$ & $0.041\pm 0.029$ & $0.018\pm 0.011$ \\
GPT-Image-1 & $0.002\pm 0.002$ & $0.023\pm 0.019$ & $0.006\pm 0.007$ & $0.004\pm 0.004$ & $0.028\pm 0.023$ & $0.048\pm 0.041$ \\
SD 3.5 & $0.006\pm 0.011$ & $0.017\pm 0.011$ & $0.005\pm 0.004$ & $0.001\pm 0.0$ & $0.027\pm 0.022$ & $0.024\pm 0.013$ \\
\bottomrule
\end{tabular}

\begin{tabular}{|c|c|c|c|c|c|c|}
\toprule
Class & \multicolumn{2}{c|}{waffle\_iron} & wooden\_spoon & \multicolumn{2}{c|}{zebra} \\
\midrule
Concept & metal & waffled & wood & striped & tattoo \\
\midrule
Real & $0.052\pm 0.034$ & $0.098\pm 0.059$ & $0.033\pm 0.032$ & $0.195\pm 0.148$ & $0.028\pm 0.026$ \\
Flux & $0.027\pm 0.02$ & $0.06\pm 0.044$ & $0.025\pm 0.02$ & $0.111\pm 0.076$ & $0.012\pm 0.009$ \\
GPT-Image-1 & $0.011\pm 0.011$ & $0.061\pm 0.042$ & $0.025\pm 0.026$ & $0.089\pm 0.08$ & $0.009\pm 0.009$ \\
SD 3.5 & $0.024\pm 0.022$ & $0.095\pm 0.041$ & $0.014\pm 0.011$ & $0.033\pm 0.035$ & $0.04\pm 0.043$ \\
\bottomrule
\end{tabular}

\label{tab:concept_importances}
\caption{Average importance scores for each concept computed with Visual-TCAV. Results are aggregated by averaging across four architectures and two layers each (i.e., the outputs of the final two blocks for ResNet-50-v2, Inception-V3, and ConvNeXt, and the outputs of the last two convolutional layers of VGG-16).}
\end{table}

\vspace*{\fill}

\newpage

\section{Average Importances per Concept using TCAV}
\label{sec:app_importances_tcav}

\begin{table}[h!]
\footnotesize
\centering
\begin{tabular}{|c|c|c|c|c|c|c|}
\toprule
Class & \multicolumn{3}{c|}{beer\_glass} & \multicolumn{2}{c|}{church} & crossword\_puzzle \\
\midrule
Concept & bubbly & glass & water & cloister & steeple & chequered \\
\midrule
Real & $0.72\pm 0.301$ & $0.872\pm 0.185$ & $0.529\pm 0.452$ & $0.789\pm 0.301$ & $0.995\pm 0.011$ & $0.638\pm 0.401$ \\
Flux & $0.635\pm 0.486$ & $0.858\pm 0.293$ & $0.676\pm 0.429$ & $0.929\pm 0.137$ & $0.982\pm 0.036$ & $0.81\pm 0.22$ \\
GPT-Image-1 & $0.789\pm 0.323$ & $0.978\pm 0.044$ & $0.122\pm 0.144$ & $0.71\pm 0.347$ & $0.979\pm 0.026$ & $0.65\pm 0.413$ \\
SD 3.5 & $0.721\pm 0.419$ & $0.654\pm 0.407$ & $0.397\pm 0.315$ & $0.84\pm 0.22$ & $0.985\pm 0.039$ & $0.678\pm 0.419$ \\
\bottomrule
\end{tabular}

\centering
\begin{tabular}{|c|c|c|c|c|c|c|}
\toprule
Class & \multicolumn{2}{c|}{crossword\_puzzle} & \multicolumn{2}{c|}{dalmatian} & electric\_guitar & fountain\_pen \\
\midrule
Concept & newspaper & paper & dotted & spotted & guitarist & pen \\
\midrule
Real & $0.863\pm 0.223$ & $0.638\pm 0.416$ & $0.166\pm 0.197$ & $0.939\pm 0.087$ & $0.99\pm 0.015$ & $0.825\pm 0.209$ \\
Flux & $0.715\pm 0.409$ & $0.678\pm 0.409$ & $0.145\pm 0.164$ & $0.921\pm 0.091$ & $0.983\pm 0.035$ & $0.89\pm 0.148$ \\
GPT-Image-1 & $0.764\pm 0.424$ & $0.702\pm 0.301$ & $0.282\pm 0.363$ & $0.42\pm 0.387$ & $0.948\pm 0.089$ & $0.988\pm 0.028$ \\
SD 3.5 & $0.716\pm 0.433$ & $0.922\pm 0.174$ & $0.007\pm 0.02$ & $0.387\pm 0.401$ & $0.99\pm 0.021$ & $0.828\pm 0.111$ \\
\bottomrule
\end{tabular}

\centering
\begin{tabular}{|c|c|c|c|c|c|c|}
\toprule
Class & \multicolumn{3}{c|}{golf\_ball} & \multicolumn{3}{c|}{honeycomb} \\
\midrule
Concept & grass & plastic & sphere & crystalline & honeycombed & meshed \\
\midrule
Real & $0.398\pm 0.422$ & $0.311\pm 0.36$ & $0.714\pm 0.26$ & $0.61\pm 0.407$ & $0.972\pm 0.037$ & $0.867\pm 0.096$ \\
Flux & $0.368\pm 0.313$ & $0.45\pm 0.414$ & $0.735\pm 0.454$ & $0.9\pm 0.124$ & $0.968\pm 0.041$ & $0.67\pm 0.396$ \\
GPT-Image-1 & $0.881\pm 0.166$ & $0.508\pm 0.379$ & $0.706\pm 0.442$ & $0.71\pm 0.285$ & $0.972\pm 0.037$ & $0.72\pm 0.387$ \\
SD 3.5 & $0.428\pm 0.328$ & $0.636\pm 0.339$ & $0.833\pm 0.265$ & $0.645\pm 0.45$ & $0.983\pm 0.027$ & $0.802\pm 0.205$ \\
\bottomrule
\end{tabular}

\centering
\begin{tabular}{|c|c|c|c|c|c|c|}
\toprule
Class & honeycomb & \multicolumn{3}{c|}{leopard} & \multicolumn{2}{c|}{overskirt} \\
\midrule
Concept & perforated & feline & leopard\_print & rodent & frilly & woven \\
\midrule
Real & $0.698\pm 0.357$ & $0.968\pm 0.084$ & $0.977\pm 0.023$ & $0.78\pm 0.379$ & $0.474\pm 0.18$ & $0.048\pm 0.061$ \\
Flux & $0.73\pm 0.44$ & $0.948\pm 0.104$ & $0.5\pm 0.367$ & $0.562\pm 0.476$ & $0.959\pm 0.078$ & $0.628\pm 0.488$ \\
GPT-Image-1 & $0.727\pm 0.337$ & $0.935\pm 0.184$ & $0.97\pm 0.039$ & $0.878\pm 0.293$ & $0.513\pm 0.494$ & $0.61\pm 0.506$ \\
SD 3.5 & $0.695\pm 0.422$ & $0.942\pm 0.147$ & $0.668\pm 0.421$ & $0.893\pm 0.252$ & $0.98\pm 0.05$ & $0.054\pm 0.058$ \\
\bottomrule
\end{tabular}

\centering
\begin{tabular}{|c|c|c|c|c|c|c|}
\toprule
Class & overskirt & spider\_web & \multicolumn{4}{c|}{stone\_wall} \\
\midrule
Concept & wrinkled & foliage & asparagus & cracked & lichen & stone \\
\midrule
Real & $0.27\pm 0.255$ & $0.455\pm 0.421$ & $0.468\pm 0.408$ & $0.683\pm 0.429$ & $0.707\pm 0.426$ & $0.675\pm 0.411$ \\
Flux & $0.645\pm 0.418$ & $0.578\pm 0.397$ & $0.47\pm 0.46$ & $0.68\pm 0.435$ & $0.583\pm 0.453$ & $0.748\pm 0.461$ \\
GPT-Image-1 & $0.117\pm 0.182$ & $0.25\pm 0.337$ & $0.537\pm 0.395$ & $0.852\pm 0.217$ & $0.815\pm 0.299$ & $0.625\pm 0.409$ \\
SD 3.5 & $0.439\pm 0.374$ & $0.46\pm 0.383$ & $0.395\pm 0.402$ & $0.635\pm 0.426$ & $0.708\pm 0.442$ & $0.985\pm 0.035$ \\
\bottomrule
\end{tabular}

\centering
\begin{tabular}{|c|c|c|c|c|c|c|}
\toprule
Class & \multicolumn{2}{c|}{tench} & \multicolumn{2}{c|}{trench\_coat} & \multicolumn{2}{c|}{waffle\_iron} \\
\midrule
Concept & fin & grandfather & fabric & leather & cast\_iron & kitchen \\
\midrule
Real & $0.572\pm 0.406$ & $0.502\pm 0.382$ & $0.4\pm 0.252$ & $0.375\pm 0.12$ & $0.74\pm 0.231$ & $0.805\pm 0.176$ \\
Flux & $0.467\pm 0.239$ & $0.535\pm 0.414$ & $0.518\pm 0.405$ & $0.17\pm 0.064$ & $0.938\pm 0.073$ & $0.867\pm 0.133$ \\
GPT-Image-1 & $0.575\pm 0.332$ & $0.565\pm 0.356$ & $0.552\pm 0.344$ & $0.222\pm 0.332$ & $0.84\pm 0.306$ & $0.942\pm 0.074$ \\
SD 3.5 & $0.7\pm 0.331$ & $0.553\pm 0.381$ & $0.32\pm 0.245$ & $0.04\pm 0.037$ & $0.68\pm 0.433$ & $0.93\pm 0.109$ \\
\bottomrule
\end{tabular}

\centering
\begin{tabular}{|c|c|c|c|c|c|}
\toprule
Class & \multicolumn{2}{c|}{waffle\_iron} & wooden\_spoon & \multicolumn{2}{c|}{zebra} \\
\midrule
Concept & metal & waffled & wood & striped & tattoo \\
\midrule
Real & $0.69\pm 0.3$ & $0.705\pm 0.269$ & $0.368\pm 0.271$ & $0.925\pm 0.101$ & $0.347\pm 0.374$ \\
Flux & $0.685\pm 0.435$ & $0.832\pm 0.224$ & $0.718\pm 0.424$ & $0.993\pm 0.01$ & $0.527\pm 0.4$ \\
GPT-Image-1 & $0.49\pm 0.392$ & $0.785\pm 0.23$ & $0.5\pm 0.382$ & $0.9\pm 0.115$ & $0.465\pm 0.344$ \\
SD 3.5 & $0.513\pm 0.356$ & $0.84\pm 0.204$ & $0.55\pm 0.41$ & $0.617\pm 0.396$ & $0.415\pm 0.398$ \\
\bottomrule
\end{tabular}

\label{tab:tcav_importances_all}
\caption{Average importance scores for each concept computed with TCAV. Results are aggregated by averaging across four architectures and two layers each (i.e., the outputs of the final two blocks for ResNet-50-v2, Inception-V3, and ConvNeXt, and the outputs of the last two convolutional layers of VGG-16).}
\end{table}

\vspace*{\fill}
\newpage

\section{Generated Concepts}
\label{sec:app_gen_concepts}

Below is a randomly selected subset of the synthetic concepts generated in this study.

\vspace*{\fill}

\subsection{Glass Concept}

\begin{figure}[h!]
    \centering
    \subfloat[Real glass concept images from FMD.]{%
        \includegraphics[width=0.19\columnwidth]{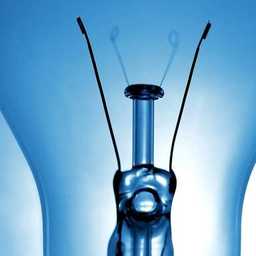}
        \includegraphics[width=0.19\columnwidth]{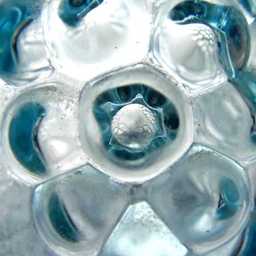}
        \includegraphics[width=0.19\columnwidth]{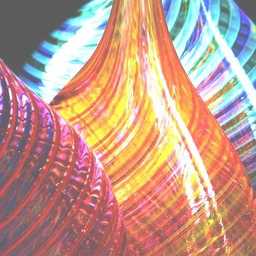}
        \includegraphics[width=0.19\columnwidth]{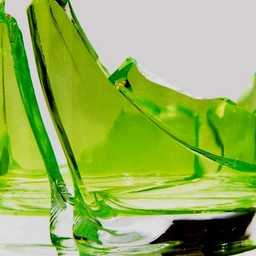}
        \includegraphics[width=0.19\columnwidth]{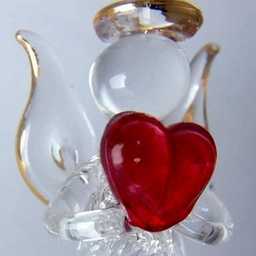}
    }
    \quad
    \subfloat[Glass concept images generated with Flux 1.1.]{%
        \includegraphics[width=0.19\columnwidth]{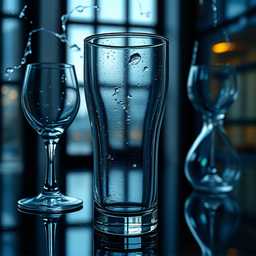}
        \includegraphics[width=0.19\columnwidth]{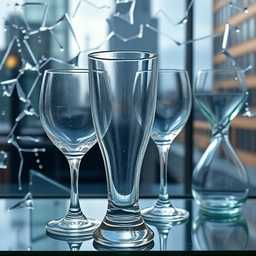}
        \includegraphics[width=0.19\columnwidth]{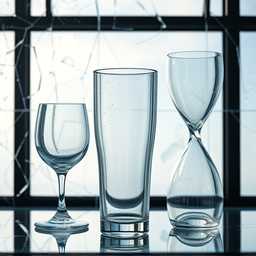}
        \includegraphics[width=0.19\columnwidth]{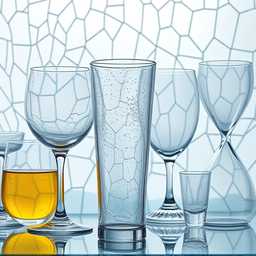}
        \includegraphics[width=0.19\columnwidth]{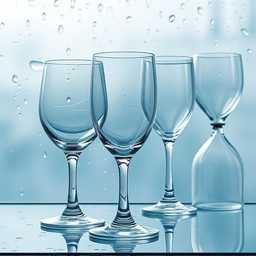}
    }
    \quad
    \subfloat[Glass concept images generated with GPT-Image-1.]{%
        \includegraphics[width=0.19\columnwidth]{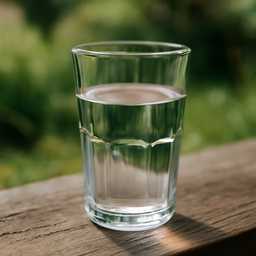}
        \includegraphics[width=0.19\columnwidth]{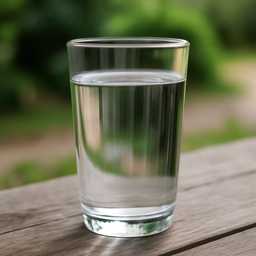}
        \includegraphics[width=0.19\columnwidth]{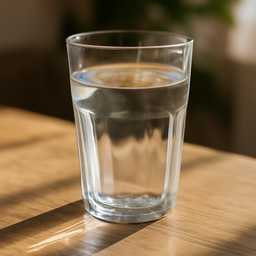}
        \includegraphics[width=0.19\columnwidth]{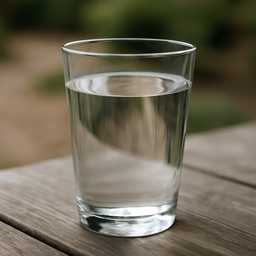}
        \includegraphics[width=0.19\columnwidth]{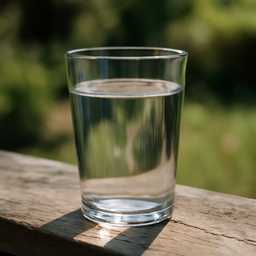}
    }
    \quad
    \subfloat[Glass concept images generated with Stable Diffusion 3.5 Medium.]{%
        \includegraphics[width=0.19\columnwidth]{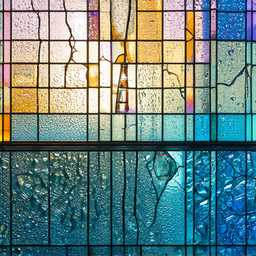}
        \includegraphics[width=0.19\columnwidth]{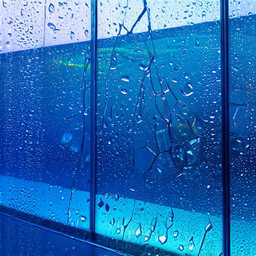}
        \includegraphics[width=0.19\columnwidth]{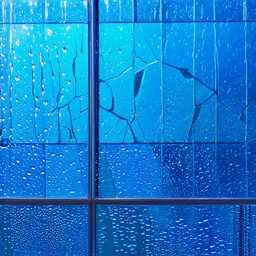}
        \includegraphics[width=0.19\columnwidth]{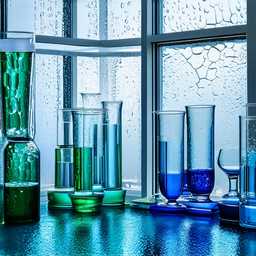}
        \includegraphics[width=0.19\columnwidth]{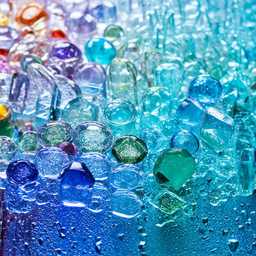}
    }
    \caption{Comparison of real glass images and generated concept images across various models.}
\end{figure}

\vspace*{\fill}

\newpage

\vspace*{\fill}

\subsection{Water Concept}

\begin{figure}[h!]
    \centering
    \subfloat[Real water concept images from FMD.]{%
        \includegraphics[width=0.19\columnwidth]{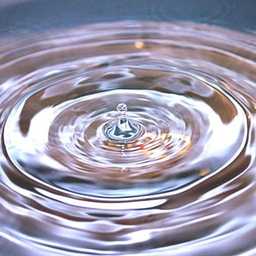}
        \includegraphics[width=0.19\columnwidth]{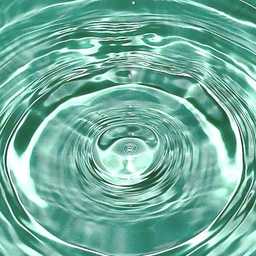}
        \includegraphics[width=0.19\columnwidth]{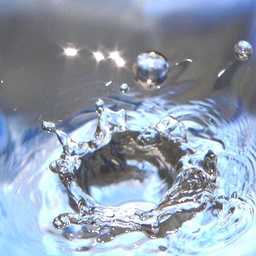}
        \includegraphics[width=0.19\columnwidth]{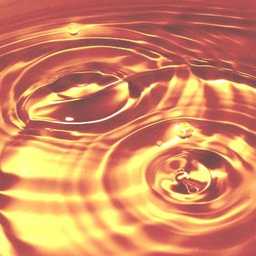}
        \includegraphics[width=0.19\columnwidth]{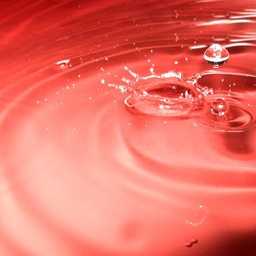}
    }
    \quad
    \subfloat[Water concept images generated with Flux 1.1.]{%
        \includegraphics[width=0.19\columnwidth]{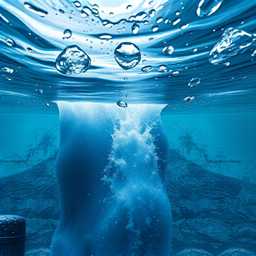}
        \includegraphics[width=0.19\columnwidth]{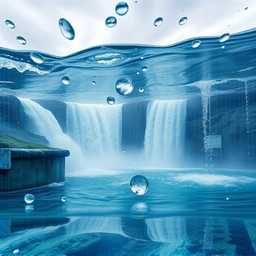}
        \includegraphics[width=0.19\columnwidth]{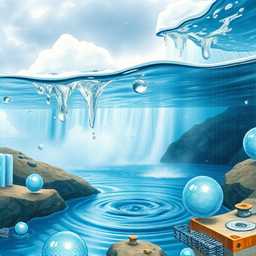}
        \includegraphics[width=0.19\columnwidth]{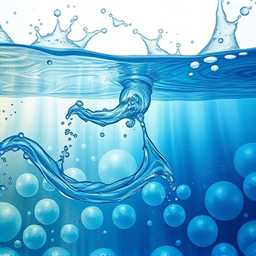}
        \includegraphics[width=0.19\columnwidth]{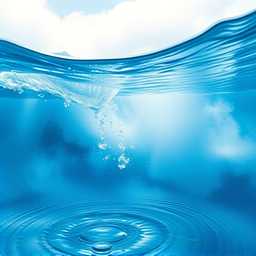}
    }
    \quad
    \subfloat[Water concept images generated with GPT-Image-1.]{%
        \includegraphics[width=0.19\columnwidth]{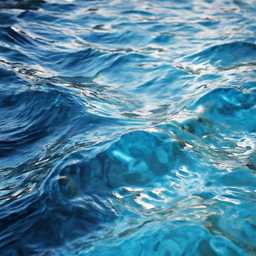}
        \includegraphics[width=0.19\columnwidth]{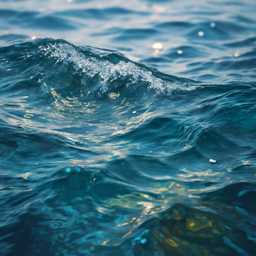}
        \includegraphics[width=0.19\columnwidth]{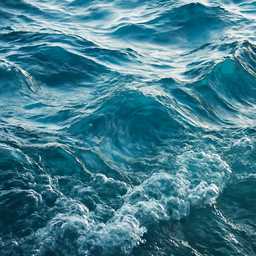}
        \includegraphics[width=0.19\columnwidth]{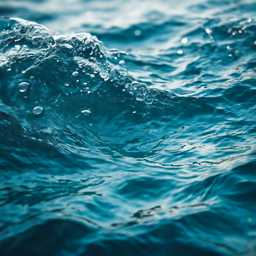}
        \includegraphics[width=0.19\columnwidth]{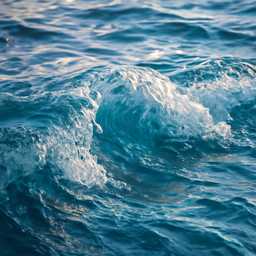}
    }
    \quad
    \subfloat[Water concept images generated with Stable Diffusion 3.5 Medium.]{%
        \includegraphics[width=0.19\columnwidth]{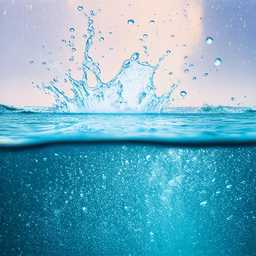}
        \includegraphics[width=0.19\columnwidth]{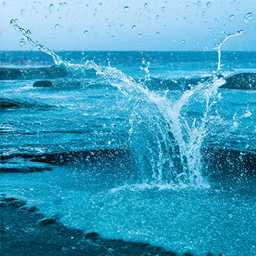}
        \includegraphics[width=0.19\columnwidth]{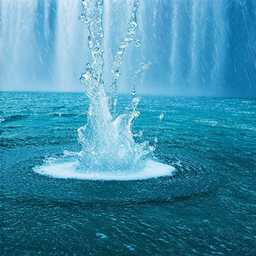}
        \includegraphics[width=0.19\columnwidth]{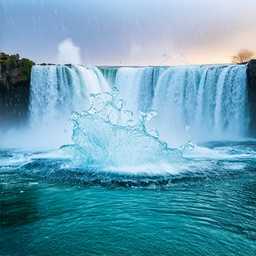}
        \includegraphics[width=0.19\columnwidth]{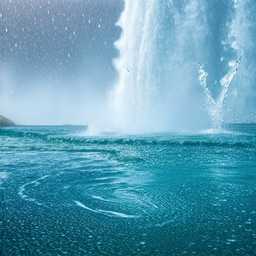}
    }
    \caption{A comparison of real images (top row) with the corresponding concept images produced by the different generative models: Flux 1.1, GPT‑Image‑1, Stable Diffusion 3.5 Medium.}
\end{figure}

\vspace*{\fill}

\newpage

\vspace*{\fill}

\subsection{Bubbly Concept}

\begin{figure}[h!]
    \centering
    \subfloat[Real bubbly concept images from DTD.]{%
        \includegraphics[width=0.19\columnwidth]{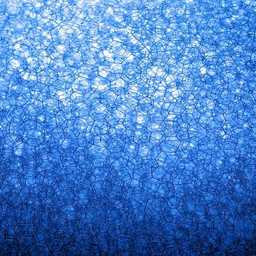}
        \includegraphics[width=0.19\columnwidth]{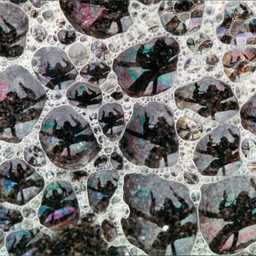}
        \includegraphics[width=0.19\columnwidth]{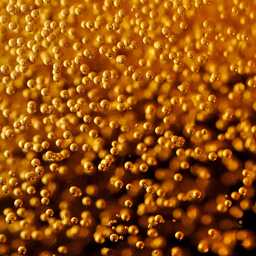}
        \includegraphics[width=0.19\columnwidth]{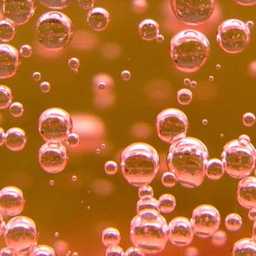}
        \includegraphics[width=0.19\columnwidth]{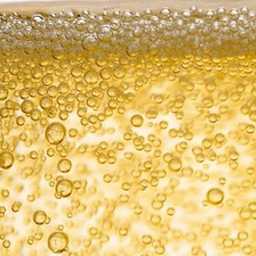}
    }
    \quad
    \subfloat[Bubbly concept images generated with Flux 1.1.]{%
        \includegraphics[width=0.19\columnwidth]{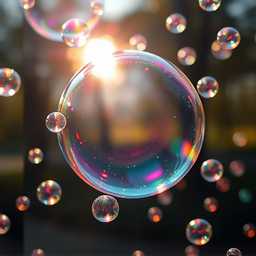}
        \includegraphics[width=0.19\columnwidth]{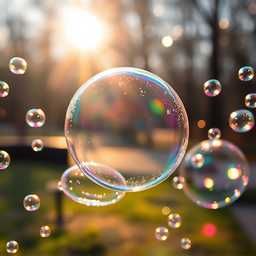}
        \includegraphics[width=0.19\columnwidth]{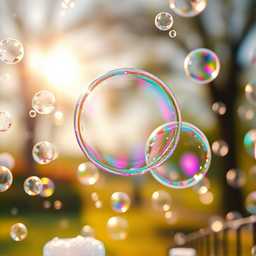}
        \includegraphics[width=0.19\columnwidth]{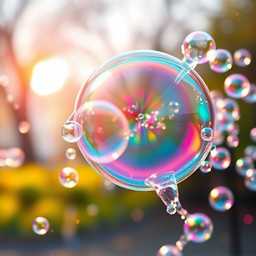}
        \includegraphics[width=0.19\columnwidth]{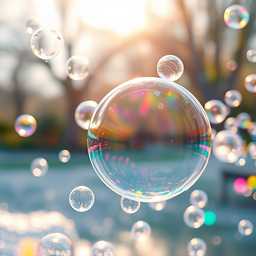}
    }
    \quad
    \subfloat[Bubbly concept images generated with GPT-Image-1.]{%
        \includegraphics[width=0.19\columnwidth]{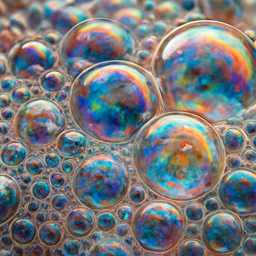}
        \includegraphics[width=0.19\columnwidth]{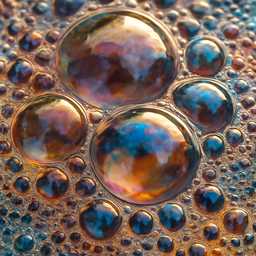}
        \includegraphics[width=0.19\columnwidth]{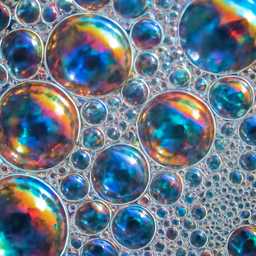}
        \includegraphics[width=0.19\columnwidth]{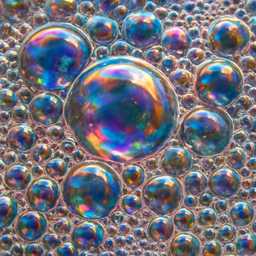}
        \includegraphics[width=0.19\columnwidth]{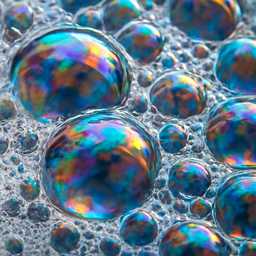}
    }
    \quad
    \subfloat[Bubbly concept images generated with Stable Diffusion 3.5 Medium.]{%
        \includegraphics[width=0.19\columnwidth]{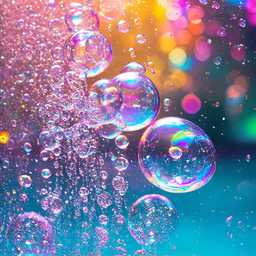}
        \includegraphics[width=0.19\columnwidth]{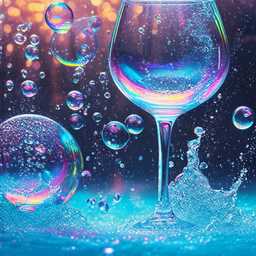}
        \includegraphics[width=0.19\columnwidth]{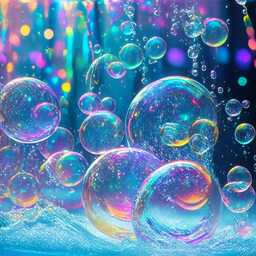}
        \includegraphics[width=0.19\columnwidth]{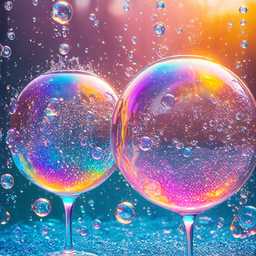}
        \includegraphics[width=0.19\columnwidth]{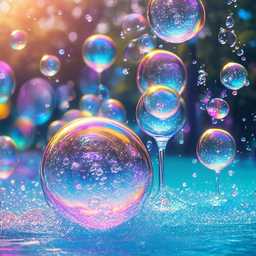}
    }
    \caption{Real bubbly images from the DTD dataset (top row) compared with concept images generated by Flux 1.1, GPT‑Image‑1, and Stable Diffusion 3.5 Medium (subsequent rows).}
\end{figure}

\vspace*{\fill}

\newpage

\vspace*{\fill}

\subsection{Cloister Concept}

\begin{figure}[h!]
    \centering
    \subfloat[Real cloister concept images from ImageNet.]{%
        \includegraphics[width=0.19\columnwidth]{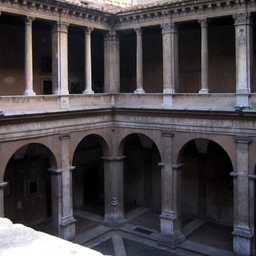}
        \includegraphics[width=0.19\columnwidth]{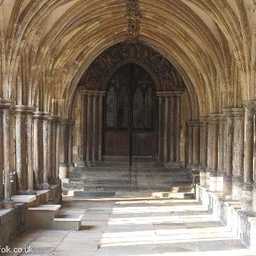}
        \includegraphics[width=0.19\columnwidth]{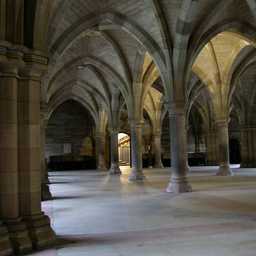}
        \includegraphics[width=0.19\columnwidth]{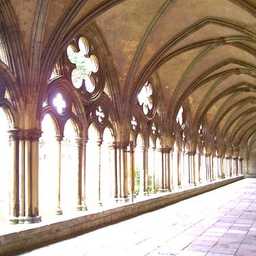}
        \includegraphics[width=0.19\columnwidth]{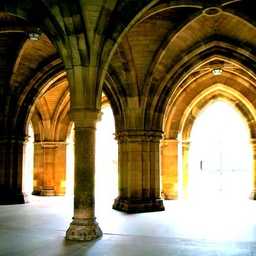}
    }
    \quad
    \subfloat[Cloister concept images generated with Flux 1.1.]{%
        \includegraphics[width=0.19\columnwidth]{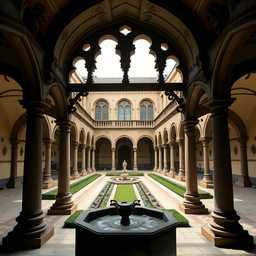}
        \includegraphics[width=0.19\columnwidth]{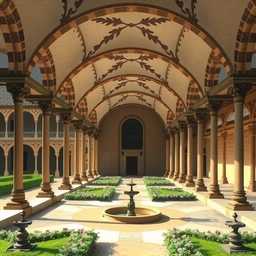}
        \includegraphics[width=0.19\columnwidth]{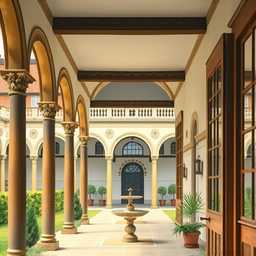}
        \includegraphics[width=0.19\columnwidth]{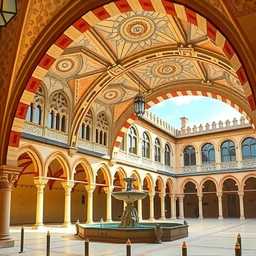}
        \includegraphics[width=0.19\columnwidth]{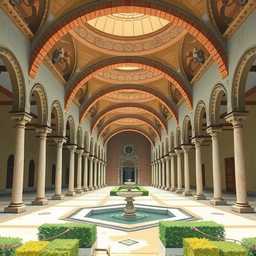}
    }
    \quad
    \subfloat[Cloister concept images generated with GPT-Image-1.]{%
        \includegraphics[width=0.19\columnwidth]{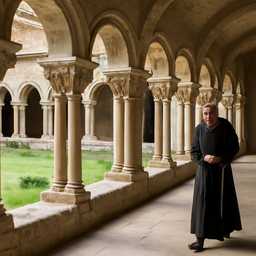}
        \includegraphics[width=0.19\columnwidth]{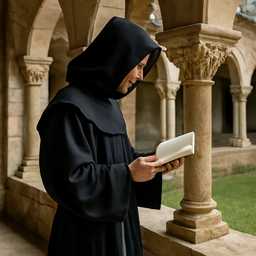}
        \includegraphics[width=0.19\columnwidth]{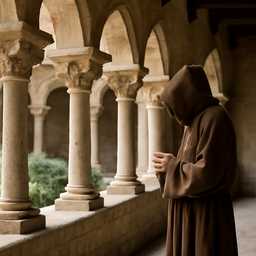}
        \includegraphics[width=0.19\columnwidth]{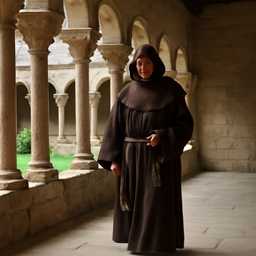}
        \includegraphics[width=0.19\columnwidth]{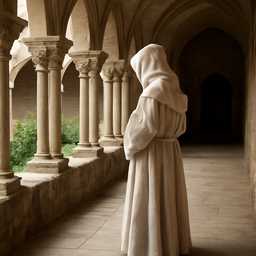}
    }
    \quad
    \subfloat[Cloister concept images generated with Stable Diffusion 3.5 Medium.]{%
        \includegraphics[width=0.19\columnwidth]{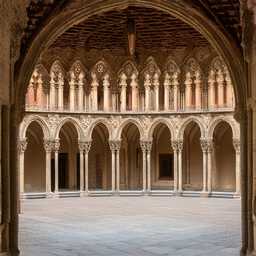}
        \includegraphics[width=0.19\columnwidth]{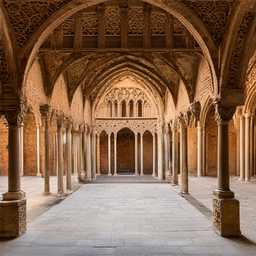}
        \includegraphics[width=0.19\columnwidth]{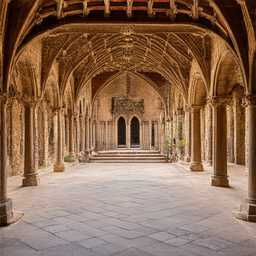}
        \includegraphics[width=0.19\columnwidth]{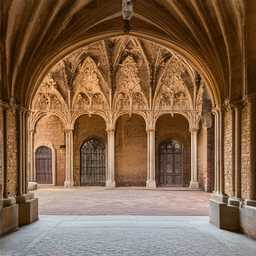}
        \includegraphics[width=0.19\columnwidth]{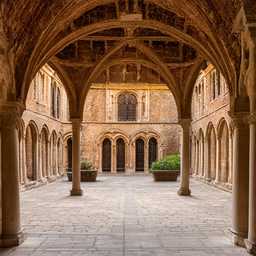}
    }
    \caption{A comparative example for the cloister concept. The first subfigure shows five authentic cloister images from the ImageNet dataset. Subsequent ones display five synthetic renditions of the same concept, generated respectively by Flux 1.1, GPT‑Image‑1, and Stable Diffusion 3.5 Medium.}
\end{figure}

\vspace*{\fill}

\newpage

\vspace*{\fill}

\subsection{Steeple Concept}

\begin{figure}[h!]
    \centering
    \subfloat[Real steeple concept images from ImageNet.]{%
        \includegraphics[width=0.19\columnwidth]{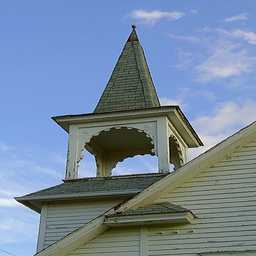}
        \includegraphics[width=0.19\columnwidth]{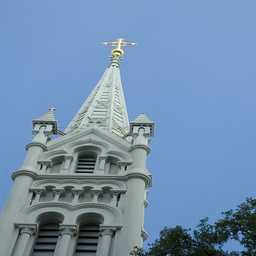}
        \includegraphics[width=0.19\columnwidth]{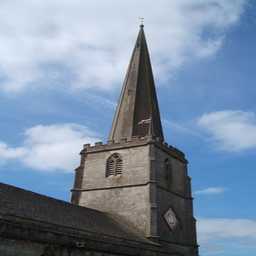}
        \includegraphics[width=0.19\columnwidth]{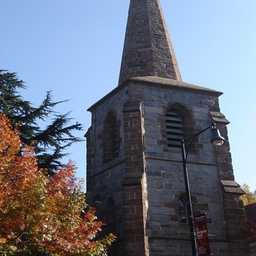}
        \includegraphics[width=0.19\columnwidth]{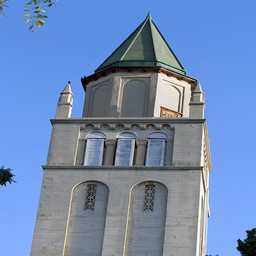}
    }
    \quad
    \subfloat[Steeple concept images generated with Flux 1.1.]{%
        \includegraphics[width=0.19\columnwidth]{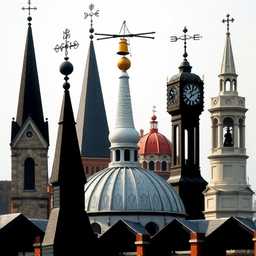}
        \includegraphics[width=0.19\columnwidth]{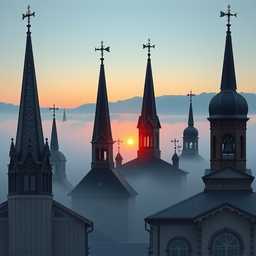}
        \includegraphics[width=0.19\columnwidth]{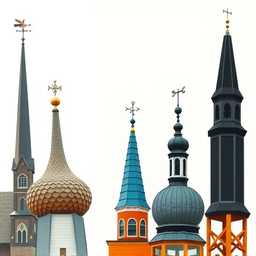}
        \includegraphics[width=0.19\columnwidth]{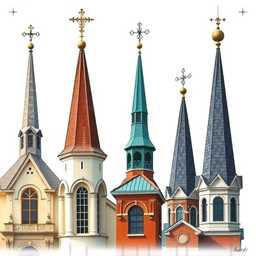}
        \includegraphics[width=0.19\columnwidth]{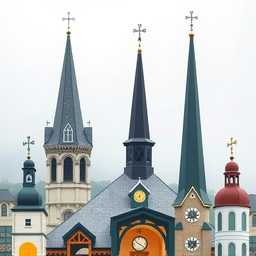}
    }
    \quad
    \subfloat[Steeple concept images generated with GPT-Image-1.]{%
        \includegraphics[width=0.19\columnwidth]{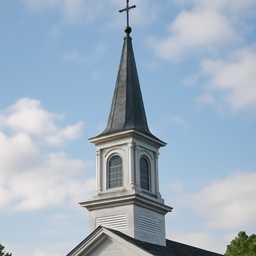}
        \includegraphics[width=0.19\columnwidth]{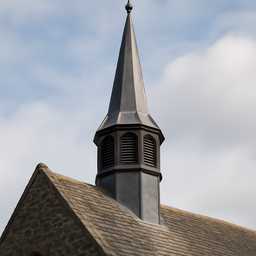}
        \includegraphics[width=0.19\columnwidth]{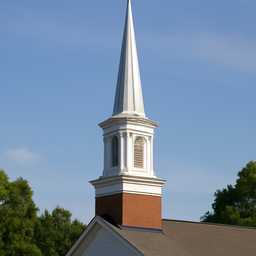}
        \includegraphics[width=0.19\columnwidth]{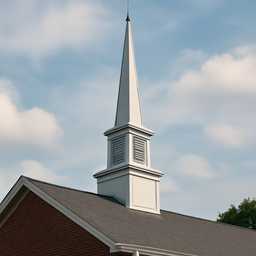}
        \includegraphics[width=0.19\columnwidth]{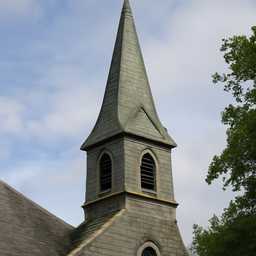}
    }
    \quad
    \subfloat[Steeple concept images generated with Stable Diffusion 3.5 Medium.]{%
        \includegraphics[width=0.19\columnwidth]{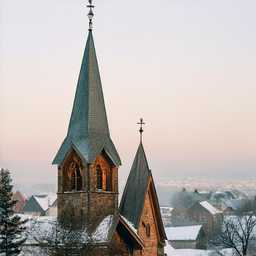}
        \includegraphics[width=0.19\columnwidth]{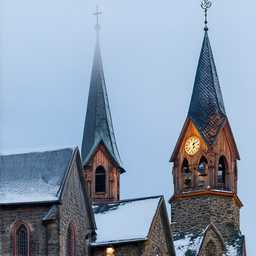}
        \includegraphics[width=0.19\columnwidth]{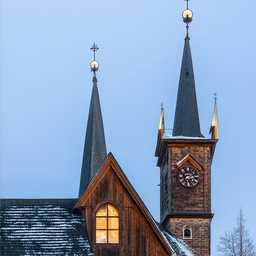}
        \includegraphics[width=0.19\columnwidth]{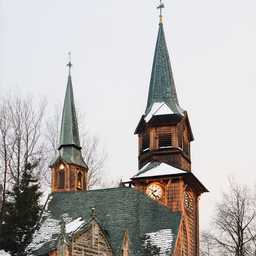}
        \includegraphics[width=0.19\columnwidth]{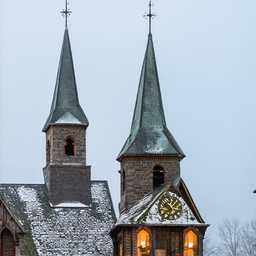}
    }
    \caption{Real steeple images from ImageNet compared with synthetic steeple images generated by Flux 1.1, GPT‑Image‑1, and Stable Diffusion 3.5 Medium. Each row contains five representative samples of the same concept.}
\end{figure}

\vspace*{\fill}

\newpage

\vspace*{\fill}

\subsection{Chequered Concept}

\begin{figure}[h!]
    \centering
    \subfloat[Real chequered concept images from DTD.]{%
        \includegraphics[width=0.19\columnwidth]{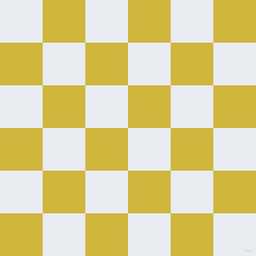}
        \includegraphics[width=0.19\columnwidth]{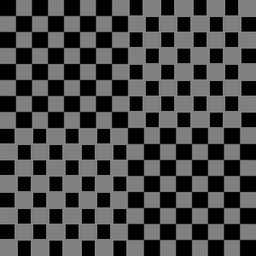}
        \includegraphics[width=0.19\columnwidth]{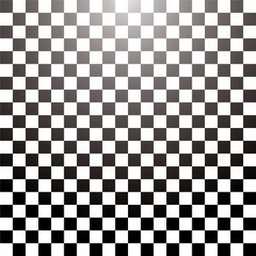}
        \includegraphics[width=0.19\columnwidth]{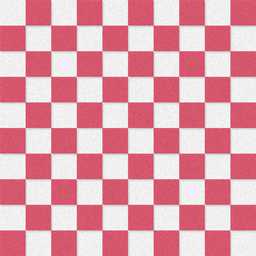}
        \includegraphics[width=0.19\columnwidth]{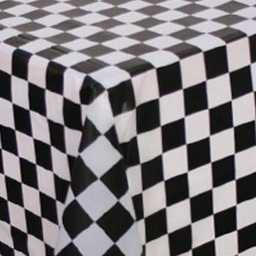}
    }
    \quad
    \subfloat[Chequered concept images generated with Flux 1.1.]{%
        \includegraphics[width=0.19\columnwidth]{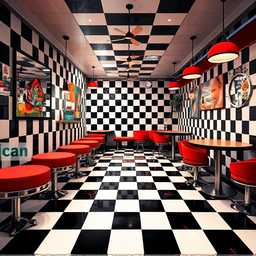}
        \includegraphics[width=0.19\columnwidth]{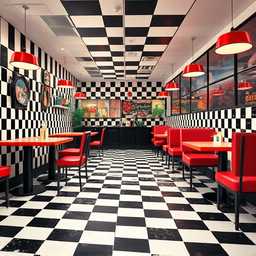}
        \includegraphics[width=0.19\columnwidth]{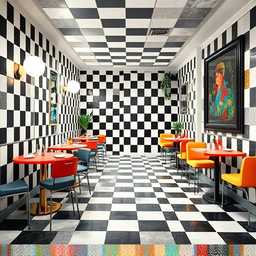}
        \includegraphics[width=0.19\columnwidth]{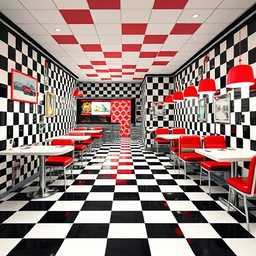}
        \includegraphics[width=0.19\columnwidth]{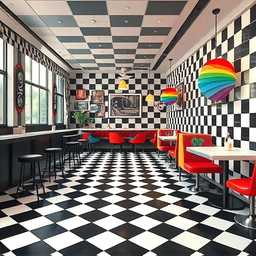}
    }
    \quad
    \subfloat[Chequered concept images generated with GPT-Image-1.]{%
        \includegraphics[width=0.19\columnwidth]{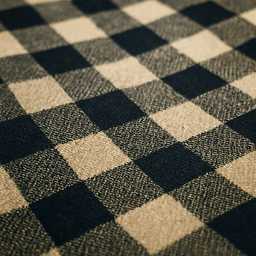}
        \includegraphics[width=0.19\columnwidth]{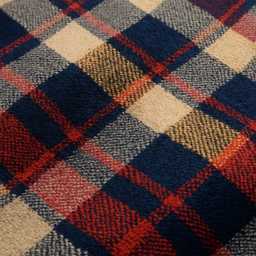}
        \includegraphics[width=0.19\columnwidth]{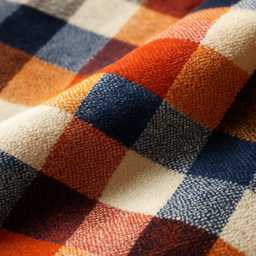}
        \includegraphics[width=0.19\columnwidth]{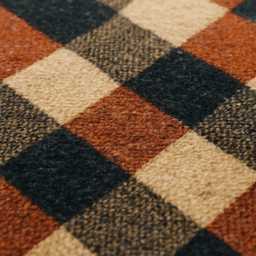}
        \includegraphics[width=0.19\columnwidth]{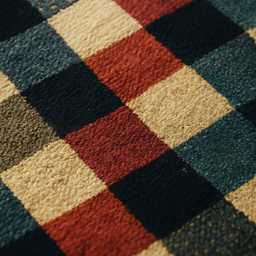}
    }
    \quad
    \subfloat[Chequered concept images generated with Stable Diffusion 3.5 Medium.]{%
        \includegraphics[width=0.19\columnwidth]{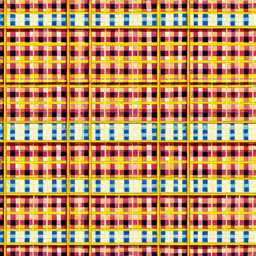}
        \includegraphics[width=0.19\columnwidth]{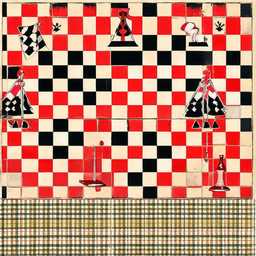}
        \includegraphics[width=0.19\columnwidth]{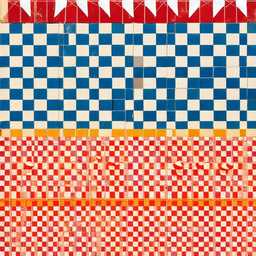}
        \includegraphics[width=0.19\columnwidth]{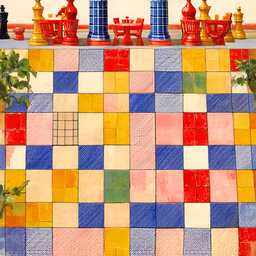}
        \includegraphics[width=0.19\columnwidth]{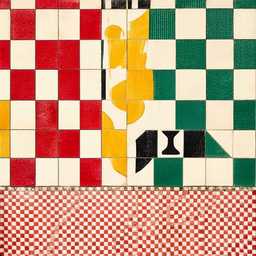}
    }
    \caption{Chequered concept images. Real DTD examples; images generated by Flux 1.1; images generated by GPT‑Image‑1; images generated by Stable Diffusion 3.5 Medium.}
\end{figure}

\vspace*{\fill}

\newpage

\vspace*{\fill}

\subsection{Newspaper Concept}

\begin{figure}[h!]
    \centering
    \subfloat[Real newspaper concept images from search engine results.]{%
        \includegraphics[width=0.19\columnwidth]{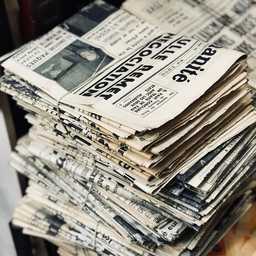}
        \includegraphics[width=0.19\columnwidth]{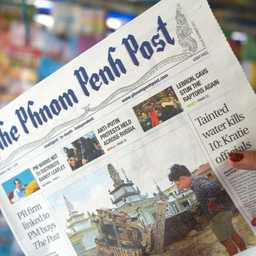}
        \includegraphics[width=0.19\columnwidth]{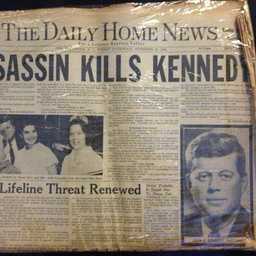}
        \includegraphics[width=0.19\columnwidth]{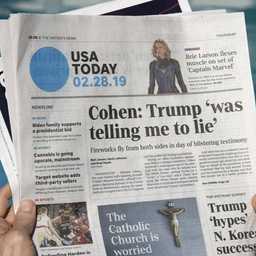}
        \includegraphics[width=0.19\columnwidth]{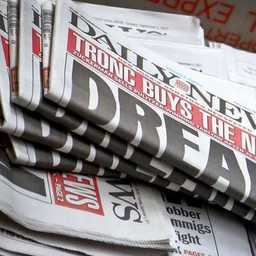}
    }
    \quad
    \subfloat[Newspaper concept images generated with Flux 1.1.]{%
        \includegraphics[width=0.19\columnwidth]{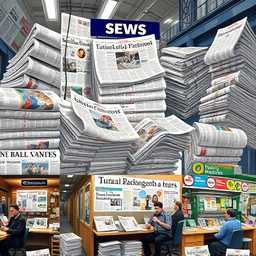}
        \includegraphics[width=0.19\columnwidth]{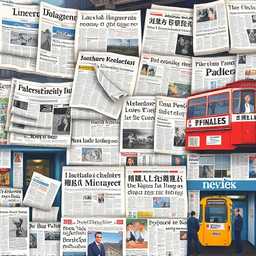}
        \includegraphics[width=0.19\columnwidth]{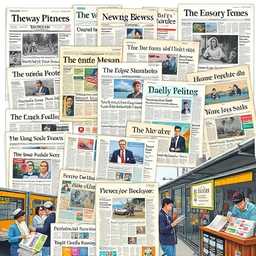}
        \includegraphics[width=0.19\columnwidth]{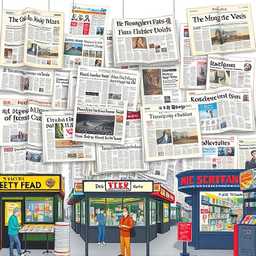}
        \includegraphics[width=0.19\columnwidth]{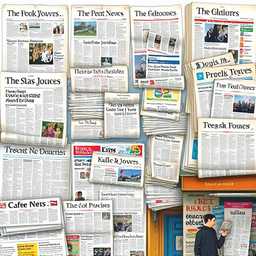}
    }
    \quad
    \subfloat[Newspaper concept images generated with GPT-Image-1.]{%
        \includegraphics[width=0.19\columnwidth]{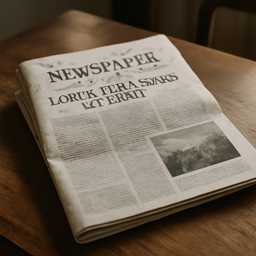}
        \includegraphics[width=0.19\columnwidth]{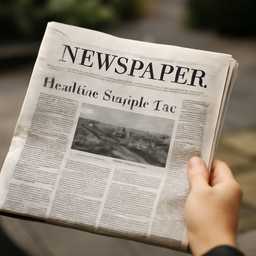}
        \includegraphics[width=0.19\columnwidth]{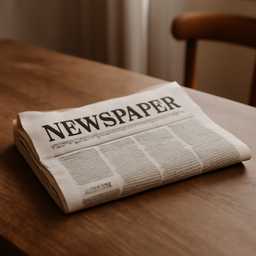}
        \includegraphics[width=0.19\columnwidth]{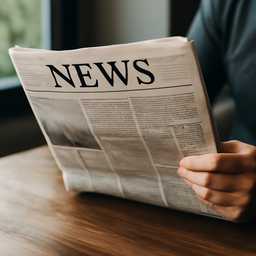}
        \includegraphics[width=0.19\columnwidth]{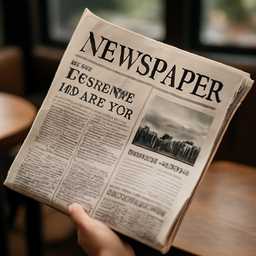}
    }
    \quad
    \subfloat[Newspaper concept images generated with Stable Diffusion 3.5 Medium.]{%
        \includegraphics[width=0.19\columnwidth]{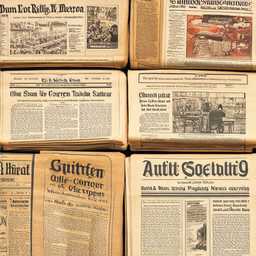}
        \includegraphics[width=0.19\columnwidth]{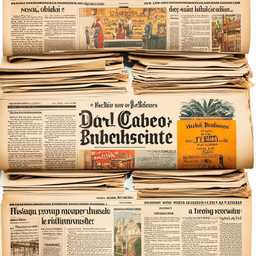}
        \includegraphics[width=0.19\columnwidth]{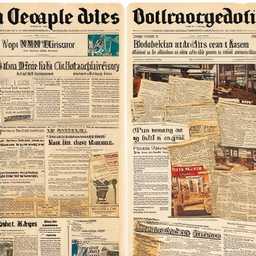}
        \includegraphics[width=0.19\columnwidth]{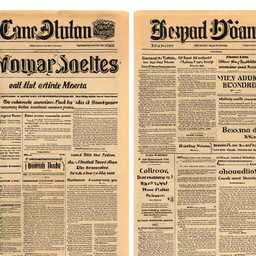}
        \includegraphics[width=0.19\columnwidth]{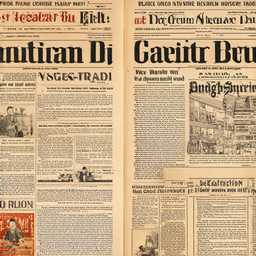}
    }
    \caption{Newspaper concept images. Row 1 shows five real samples; rows 2–4 display five synthetic images produced by Flux 1.1, GPT‑Image‑1, and Stable Diffusion 3.5 Medium, respectively.}
\end{figure}

\vspace*{\fill}

\newpage

\vspace*{\fill}

\subsection{Paper Concept}

\begin{figure}[h!]
    \centering
    \subfloat[Real paper concept images from FMD.]{%
        \includegraphics[width=0.19\columnwidth]{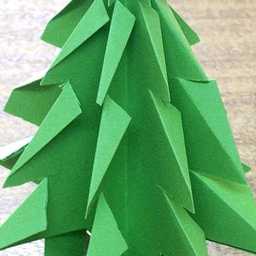}
        \includegraphics[width=0.19\columnwidth]{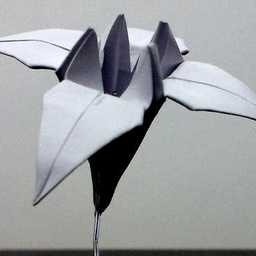}
        \includegraphics[width=0.19\columnwidth]{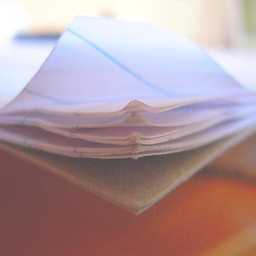}
        \includegraphics[width=0.19\columnwidth]{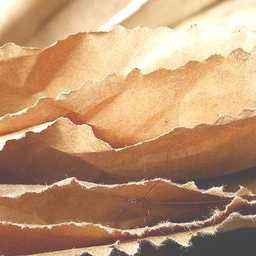}
        \includegraphics[width=0.19\columnwidth]{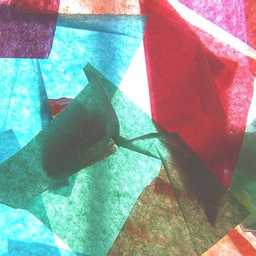}
    }
    \quad
    \subfloat[Paper concept images generated with Flux 1.1.]{%
        \includegraphics[width=0.19\columnwidth]{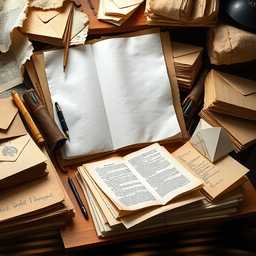}
        \includegraphics[width=0.19\columnwidth]{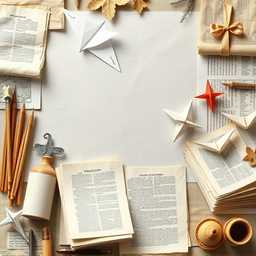}
        \includegraphics[width=0.19\columnwidth]{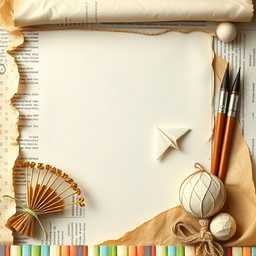}
        \includegraphics[width=0.19\columnwidth]{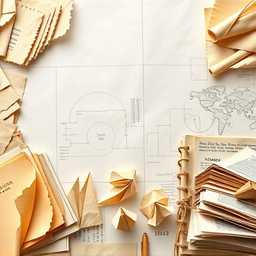}
        \includegraphics[width=0.19\columnwidth]{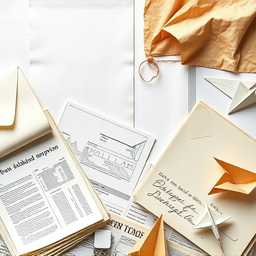}
    }
    \quad
    \subfloat[Paper concept images generated with GPT-Image-1.]{%
        \includegraphics[width=0.19\columnwidth]{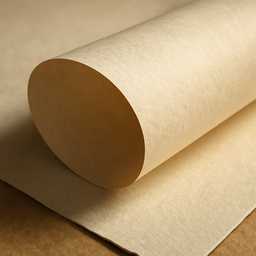}
        \includegraphics[width=0.19\columnwidth]{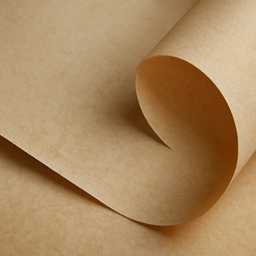}
        \includegraphics[width=0.19\columnwidth]{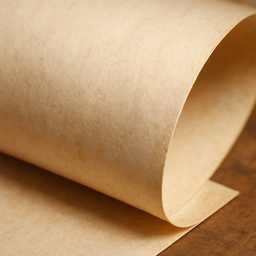}
        \includegraphics[width=0.19\columnwidth]{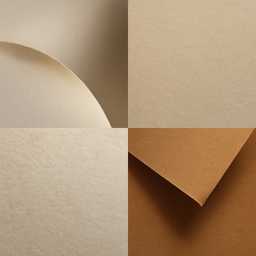}
        \includegraphics[width=0.19\columnwidth]{images/paper_example/paper_gpti1/gpti1_19.jpg}
    }
    \quad
    \subfloat[Paper concept images generated with Stable Diffusion 3.5 Medium.]{%
        \includegraphics[width=0.19\columnwidth]{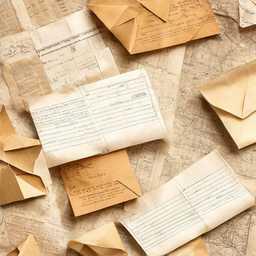}
        \includegraphics[width=0.19\columnwidth]{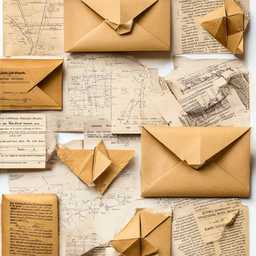}
        \includegraphics[width=0.19\columnwidth]{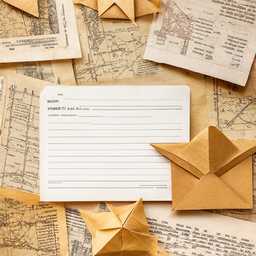}
        \includegraphics[width=0.19\columnwidth]{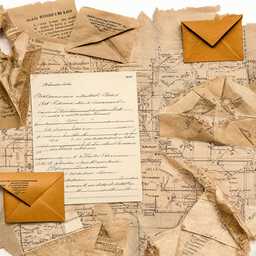}
        \includegraphics[width=0.19\columnwidth]{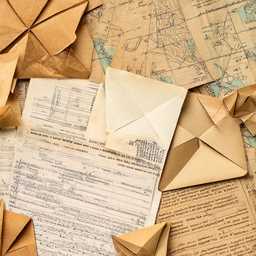}
    }
    \caption{Comparison of paper concepts: real FMD examples, Flux 1.1 generated images, GPT‑Image‑1 generated images, and Stable Diffusion 3.5 Medium generated images.}
\end{figure}

\vspace*{\fill}

\newpage

\vspace*{\fill}

\subsection{Dotted Concept}

\begin{figure}[h!]
    \centering
    \subfloat[Real dotted concept images from DTD.]{%
        \includegraphics[width=0.19\columnwidth]{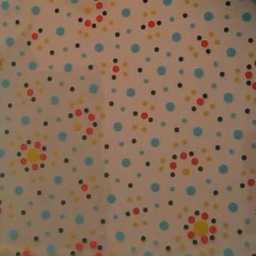}
        \includegraphics[width=0.19\columnwidth]{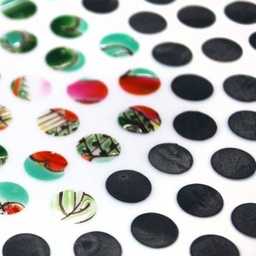}
        \includegraphics[width=0.19\columnwidth]{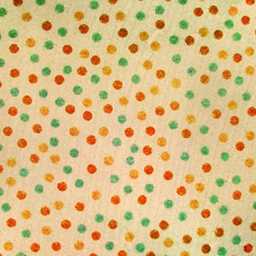}
        \includegraphics[width=0.19\columnwidth]{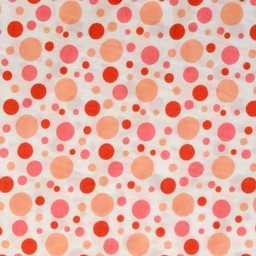}
        \includegraphics[width=0.19\columnwidth]{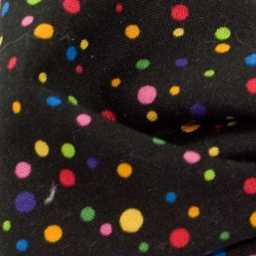}
    }
    \quad
    \subfloat[Dotted concept images generated with Flux 1.1.]{%
        \includegraphics[width=0.19\columnwidth]{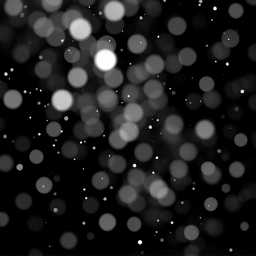}
        \includegraphics[width=0.19\columnwidth]{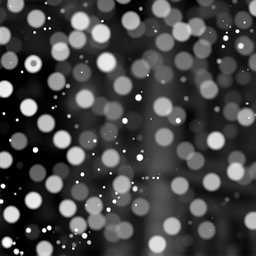}
        \includegraphics[width=0.19\columnwidth]{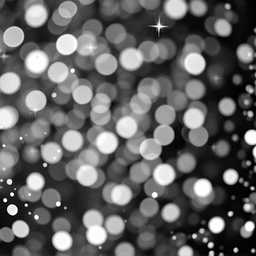}
        \includegraphics[width=0.19\columnwidth]{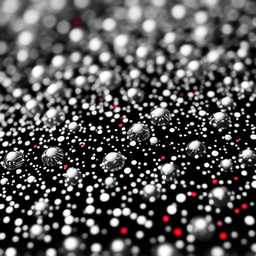}
        \includegraphics[width=0.19\columnwidth]{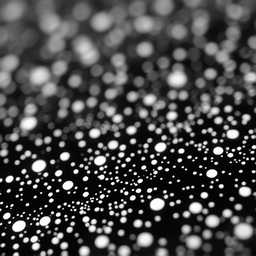}
    }
    \quad
    \subfloat[Dotted concept images generated with GPT-Image-1.]{%
        \includegraphics[width=0.19\columnwidth]{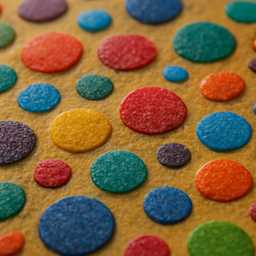}
        \includegraphics[width=0.19\columnwidth]{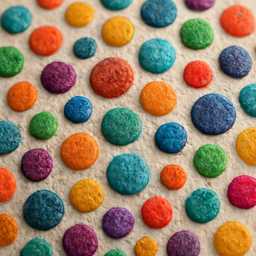}
        \includegraphics[width=0.19\columnwidth]{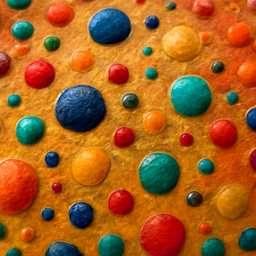}
        \includegraphics[width=0.19\columnwidth]{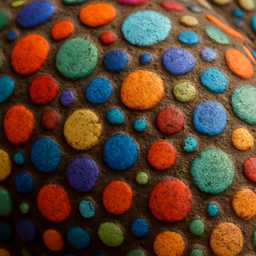}
        \includegraphics[width=0.19\columnwidth]{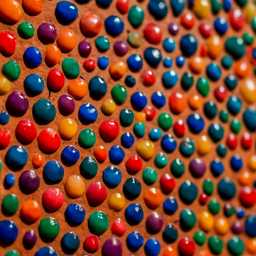}
    }
    \quad
    \subfloat[Dotted concept images generated with Stable Diffusion 3.5 Medium.]{%
        \includegraphics[width=0.19\columnwidth]{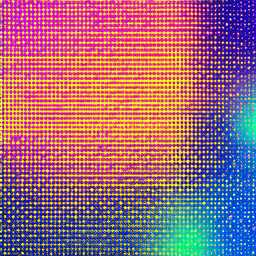}
        \includegraphics[width=0.19\columnwidth]{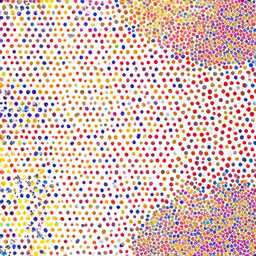}
        \includegraphics[width=0.19\columnwidth]{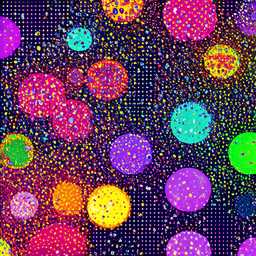}
        \includegraphics[width=0.19\columnwidth]{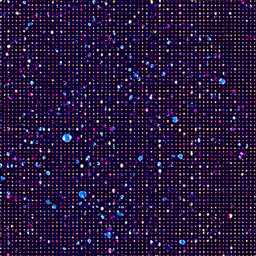}
        \includegraphics[width=0.19\columnwidth]{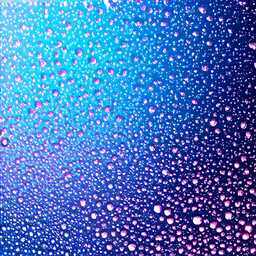}
    }
    \caption{Real dotted concept images from the DTD dataset; images produced with Flux 1.1; images produced with GPT‑Image‑1; images produced with Stable Diffusion 3.5 Medium.}
\end{figure}

\vspace*{\fill}

\newpage

\vspace*{\fill}

\subsection{Spotted Concept}

\begin{figure}[h!]
    \centering
    \subfloat[Real spotted concept images from search engine results.]{%
        \includegraphics[width=0.19\columnwidth]{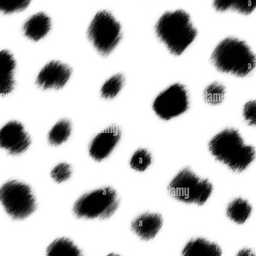}
        \includegraphics[width=0.19\columnwidth]{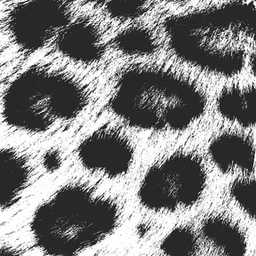}
        \includegraphics[width=0.19\columnwidth]{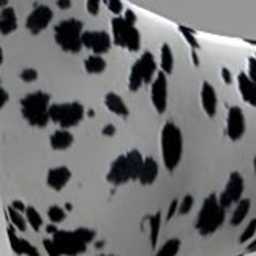}
        \includegraphics[width=0.19\columnwidth]{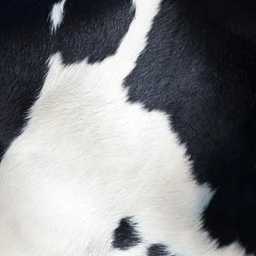}
        \includegraphics[width=0.19\columnwidth]{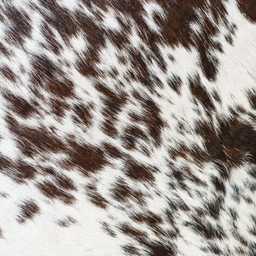}
    }
    \quad
    \subfloat[Spotted concept images generated with Flux 1.1.]{%
        \includegraphics[width=0.19\columnwidth]{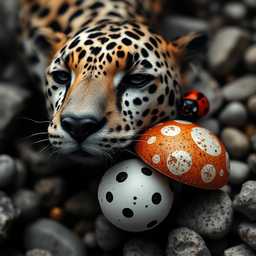}
        \includegraphics[width=0.19\columnwidth]{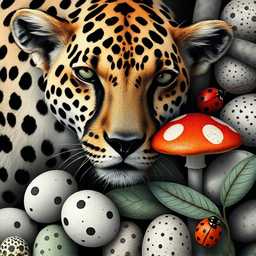}
        \includegraphics[width=0.19\columnwidth]{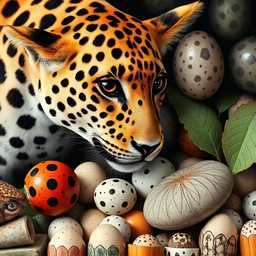}
        \includegraphics[width=0.19\columnwidth]{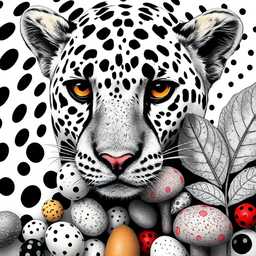}
        \includegraphics[width=0.19\columnwidth]{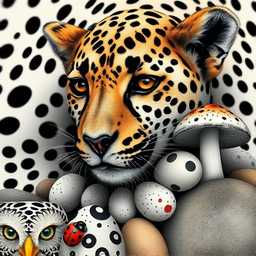}
    }
    \quad
    \subfloat[Spotted concept images generated with GPT-Image-1.]{%
        \includegraphics[width=0.19\columnwidth]{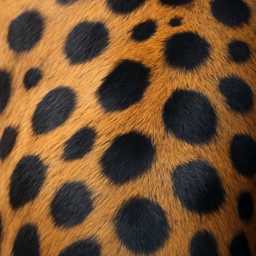}
        \includegraphics[width=0.19\columnwidth]{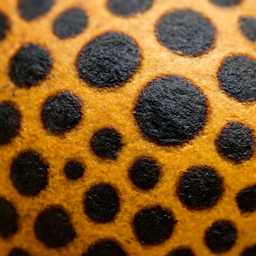}
        \includegraphics[width=0.19\columnwidth]{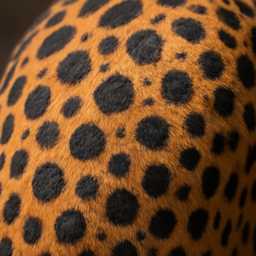}
        \includegraphics[width=0.19\columnwidth]{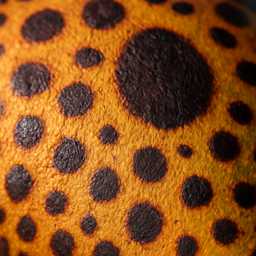}
        \includegraphics[width=0.19\columnwidth]{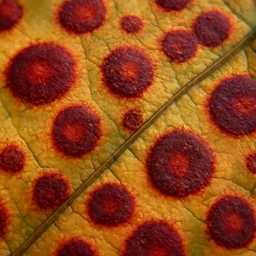}
    }
    \quad
    \subfloat[Spotted concept images generated with Stable Diffusion 3.5 Medium.]{%
        \includegraphics[width=0.19\columnwidth]{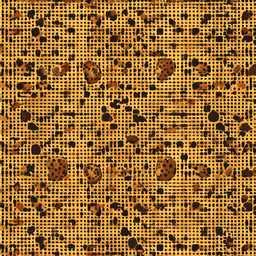}
        \includegraphics[width=0.19\columnwidth]{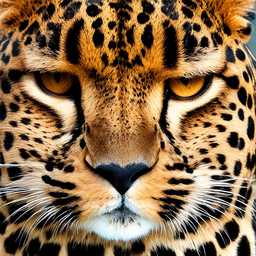}
        \includegraphics[width=0.19\columnwidth]{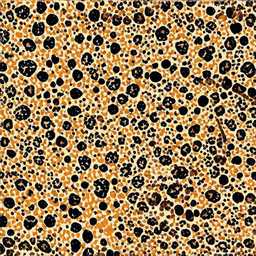}
        \includegraphics[width=0.19\columnwidth]{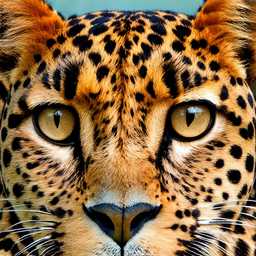}
        \includegraphics[width=0.19\columnwidth]{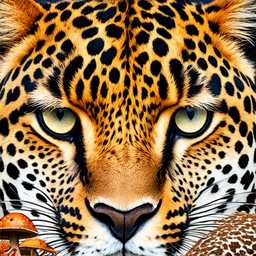}
    }
    \caption{Real and synthetic spotted‑pattern concept images.}
\end{figure}

\vspace*{\fill}

\newpage

\vspace*{\fill}

\subsection{Grass Concept}

\begin{figure}[h!]
    \centering
    \subfloat[Real grass concept images from search engine results.]{%
        \includegraphics[width=0.19\columnwidth]{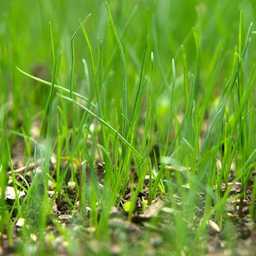}
        \includegraphics[width=0.19\columnwidth]{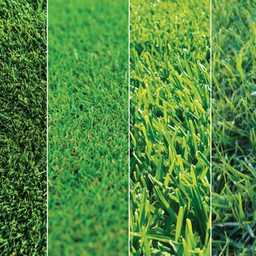}
        \includegraphics[width=0.19\columnwidth]{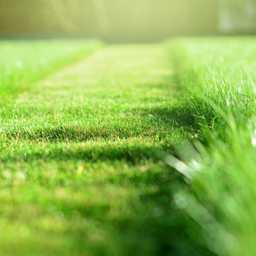}
        \includegraphics[width=0.19\columnwidth]{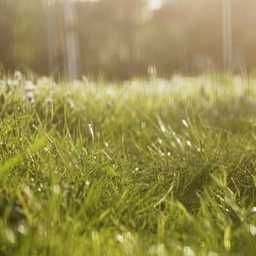}
        \includegraphics[width=0.19\columnwidth]{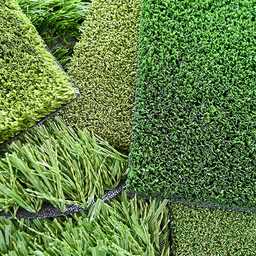}
    }
    \quad
    \subfloat[Grass concept images generated with Flux 1.1.]{%
        \includegraphics[width=0.19\columnwidth]{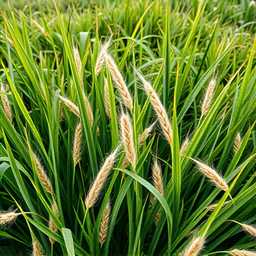}
        \includegraphics[width=0.19\columnwidth]{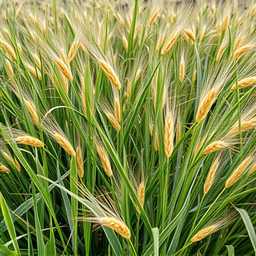}
        \includegraphics[width=0.19\columnwidth]{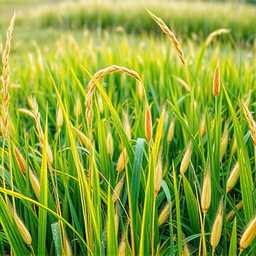}
        \includegraphics[width=0.19\columnwidth]{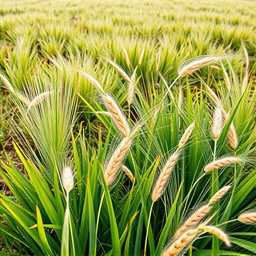}
        \includegraphics[width=0.19\columnwidth]{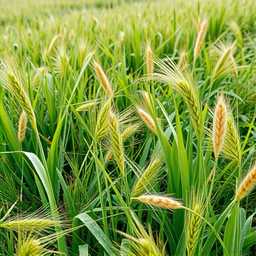}
    }
    \quad
    \subfloat[Grass concept images generated with GPT-Image-1.]{%
        \includegraphics[width=0.19\columnwidth]{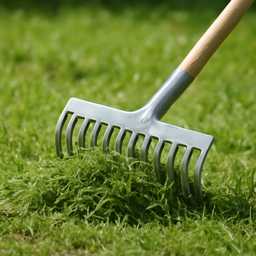}
        \includegraphics[width=0.19\columnwidth]{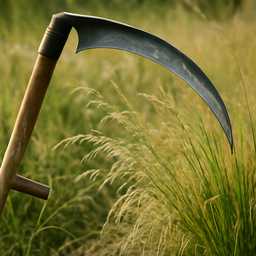}
        \includegraphics[width=0.19\columnwidth]{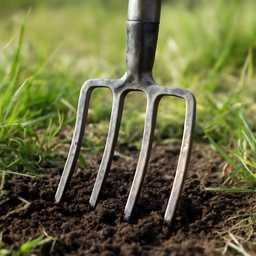}
        \includegraphics[width=0.19\columnwidth]{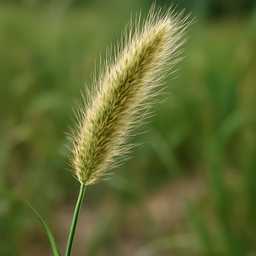}
        \includegraphics[width=0.19\columnwidth]{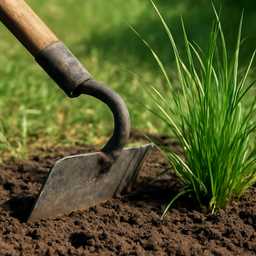}
    }
    \quad
    \subfloat[Grass concept images generated with Stable Diffusion 3.5 Medium.]{%
        \includegraphics[width=0.19\columnwidth]{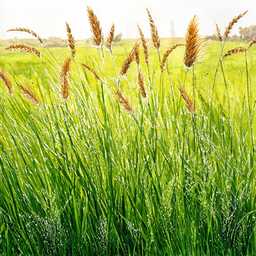}
        \includegraphics[width=0.19\columnwidth]{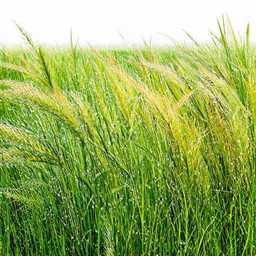}
        \includegraphics[width=0.19\columnwidth]{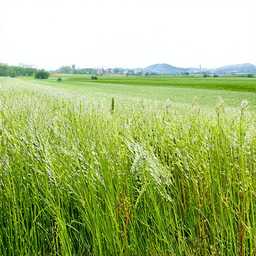}
        \includegraphics[width=0.19\columnwidth]{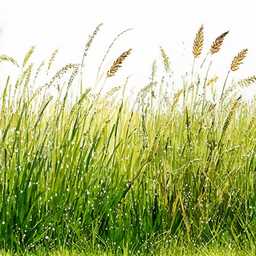}
        \includegraphics[width=0.19\columnwidth]{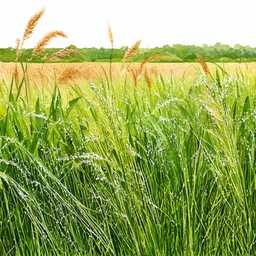}
    }
    \caption{Grass concept images: real photographs, Flux 1.1 outputs, GPT-Image-1 outputs, and Stable Diffusion 3.5 Medium outputs.}
\end{figure}

\vspace*{\fill}

\newpage

\vspace*{\fill}

\subsection{Plastic Concept}

\begin{figure}[h!]
    \centering
    \subfloat[Real plastic concept images from FMD.]{%
        \includegraphics[width=0.19\columnwidth]{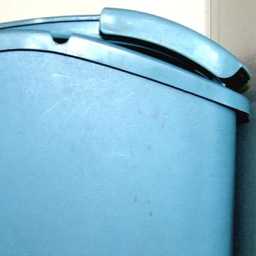}
        \includegraphics[width=0.19\columnwidth]{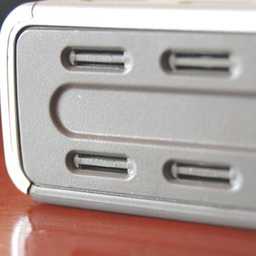}
        \includegraphics[width=0.19\columnwidth]{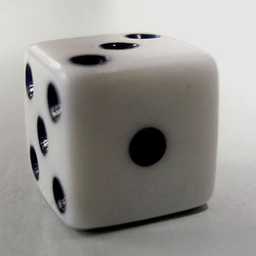}
        \includegraphics[width=0.19\columnwidth]{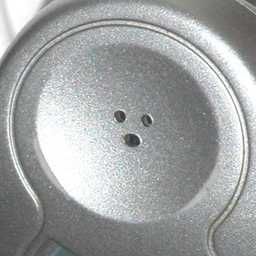}
        \includegraphics[width=0.19\columnwidth]{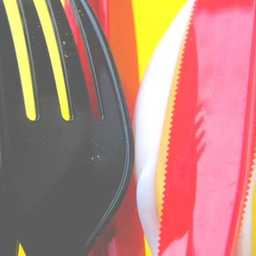}
    }
    \quad
    \subfloat[Plastic concept images generated with Flux 1.1.]{%
        \includegraphics[width=0.19\columnwidth]{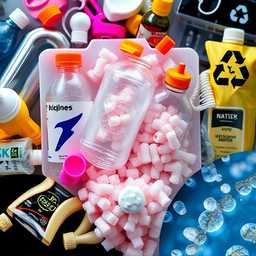}
        \includegraphics[width=0.19\columnwidth]{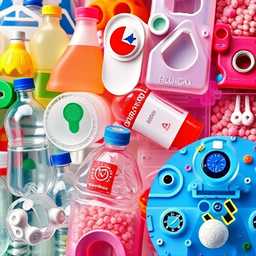}
        \includegraphics[width=0.19\columnwidth]{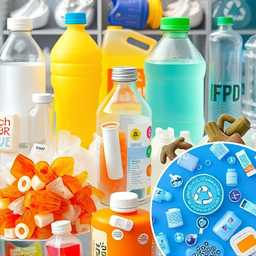}
        \includegraphics[width=0.19\columnwidth]{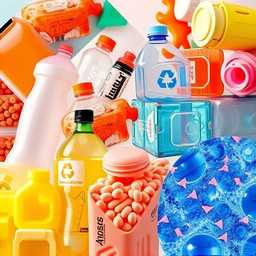}
        \includegraphics[width=0.19\columnwidth]{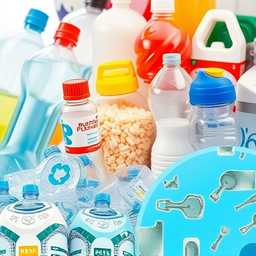}
    }
    \quad
    \subfloat[Plastic concept images generated with GPT-Image-1.]{%
        \includegraphics[width=0.19\columnwidth]{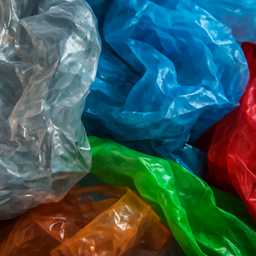}
        \includegraphics[width=0.19\columnwidth]{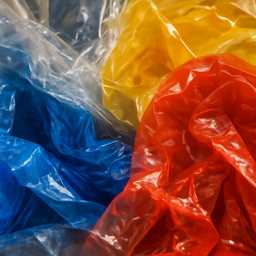}
        \includegraphics[width=0.19\columnwidth]{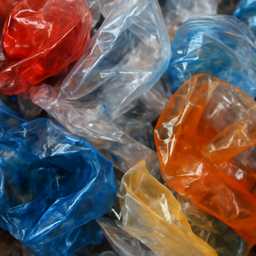}
        \includegraphics[width=0.19\columnwidth]{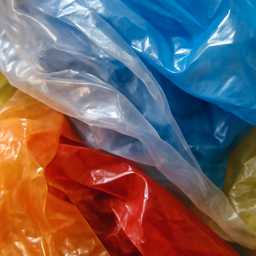}
        \includegraphics[width=0.19\columnwidth]{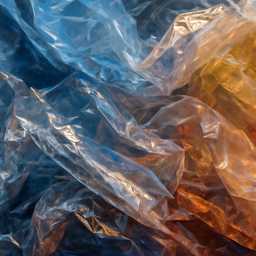}
    }
    \quad
    \subfloat[Plastic concept images generated with Stable Diffusion 3.5 Medium.]{%
        \includegraphics[width=0.19\columnwidth]{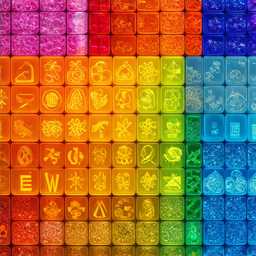}
        \includegraphics[width=0.19\columnwidth]{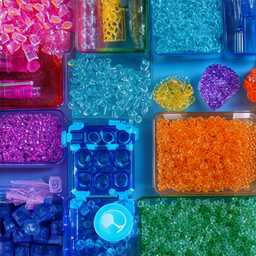}
        \includegraphics[width=0.19\columnwidth]{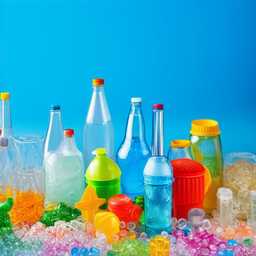}
        \includegraphics[width=0.19\columnwidth]{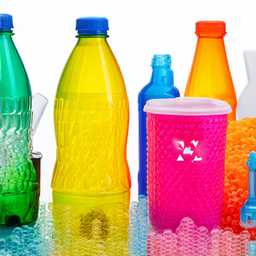}
        \includegraphics[width=0.19\columnwidth]{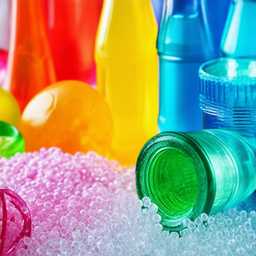}
    }
    \caption{Plastic concept images from real photographs, Flux 1.1, GPT-Image-1, and Stable Diffusion 3.5 Medium.}
\end{figure}

\vspace*{\fill}

\newpage

\vspace*{\fill}

\subsection{Sphere Concept}

\begin{figure}[h!]
    \centering
    \subfloat[Real sphere concept images from search engine results.]{%
        \includegraphics[width=0.19\columnwidth]{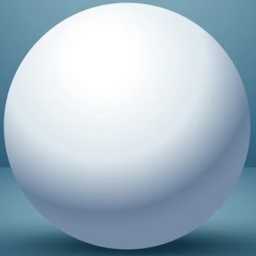}
        \includegraphics[width=0.19\columnwidth]{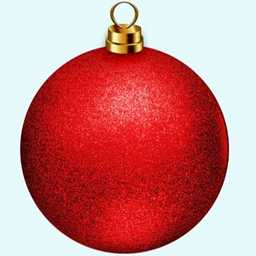}
        \includegraphics[width=0.19\columnwidth]{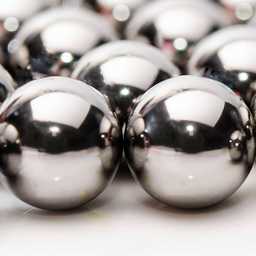}
        \includegraphics[width=0.19\columnwidth]{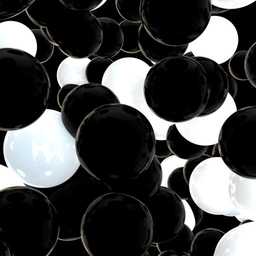}
        \includegraphics[width=0.19\columnwidth]{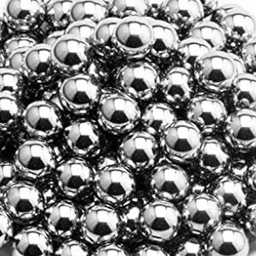}
    }
    \quad
    \subfloat[Sphere concept images generated with Flux 1.1.]{%
        \includegraphics[width=0.19\columnwidth]{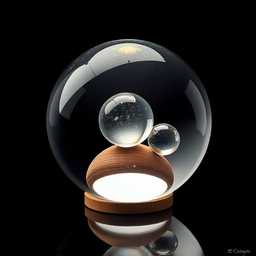}
        \includegraphics[width=0.19\columnwidth]{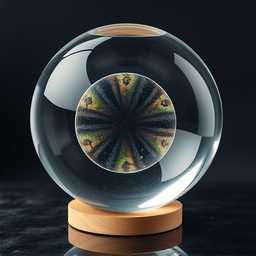}
        \includegraphics[width=0.19\columnwidth]{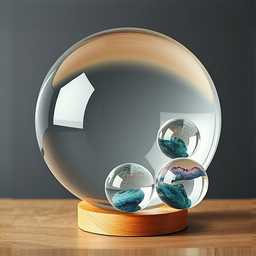}
        \includegraphics[width=0.19\columnwidth]{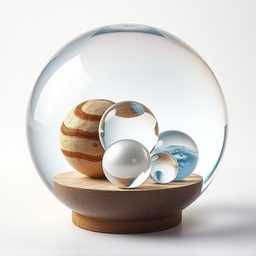}
        \includegraphics[width=0.19\columnwidth]{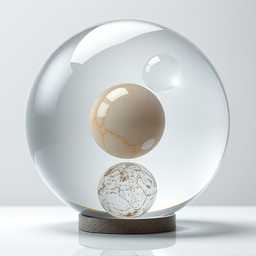}
    }
    \quad
    \subfloat[Sphere concept images generated with GPT-Image-1.]{%
        \includegraphics[width=0.19\columnwidth]{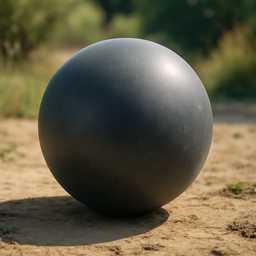}
        \includegraphics[width=0.19\columnwidth]{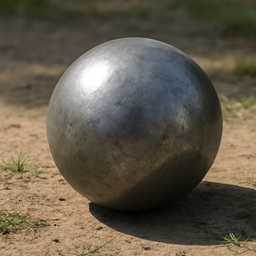}
        \includegraphics[width=0.19\columnwidth]{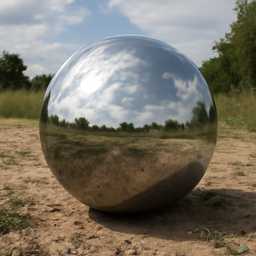}
        \includegraphics[width=0.19\columnwidth]{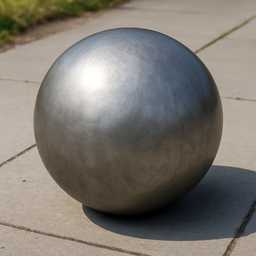}
        \includegraphics[width=0.19\columnwidth]{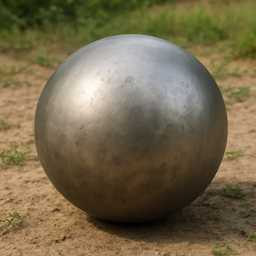}
    }
    \quad
    \subfloat[Sphere concept images generated with Stable Diffusion 3.5 Medium.]{%
        \includegraphics[width=0.19\columnwidth]{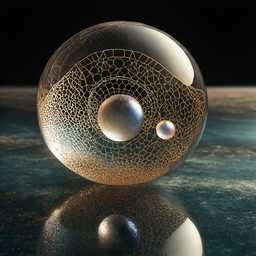}
        \includegraphics[width=0.19\columnwidth]{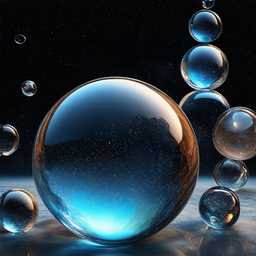}
        \includegraphics[width=0.19\columnwidth]{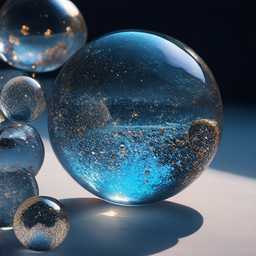}
        \includegraphics[width=0.19\columnwidth]{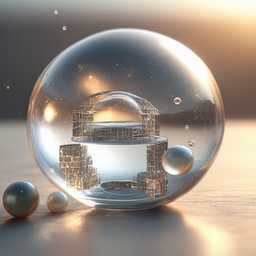}
        \includegraphics[width=0.19\columnwidth]{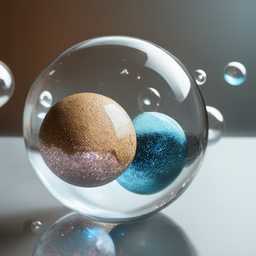}
    }
    \caption{}
\end{figure}

\vspace*{\fill}

\clearpage

\section{Ablated Class Images}
\label{sec:app_rem_images}

Below is a randomly selected subset of the ablated class images used in this study.

\vspace*{\fill}

\subsection{Beer Glass Class}

\begin{figure}[h!]
    \centering
    \subfloat[Original]{%
        \includegraphics[width=0.23\columnwidth]{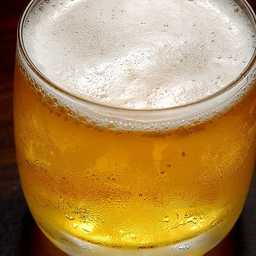}
        \label{fig:beer_glass_original}
    }
    \subfloat[No bubbly]{%
        \includegraphics[width=0.23\columnwidth]{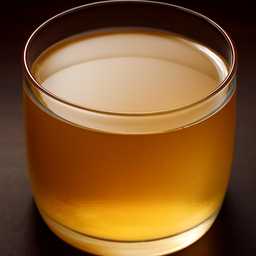}
        \label{fig:beer_glass_unbubbly}
    }
    \subfloat[No glass]{%
        \includegraphics[width=0.23\columnwidth]{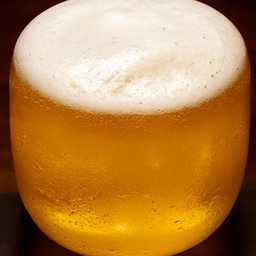}
        \label{fig:beer_glass_unglass}
    }
    \subfloat[No beer]{%
        \includegraphics[width=0.23\columnwidth]{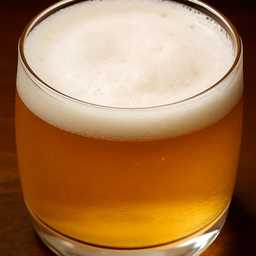}
        \label{fig:beer_glass_unwater}
    }
    \caption{Example of concept removal from a beer glass class image, eliminating bubbly, glass, and beer concepts.}
\end{figure}

\vspace*{\fill}

\subsection{Church Class}

\begin{figure}[h!]
    \centering
    \subfloat[Original]{%
        \includegraphics[width=0.25\columnwidth]{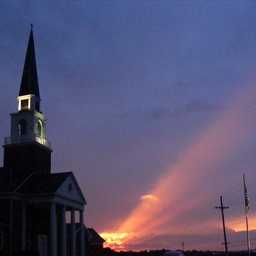}
    }
    \subfloat[No steeple]{%
        \includegraphics[width=0.25\columnwidth]{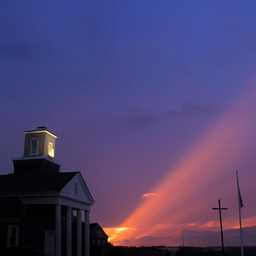}
    }
    \subfloat[No cloister]{%
        \includegraphics[width=0.25\columnwidth]{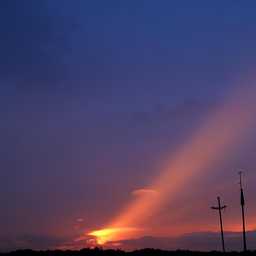}
    }
    \caption{Example of concept removal from a church class image, eliminating the steeple concept and the cloister concept.}
\end{figure}

\vspace*{\fill}

\newpage

\vspace*{\fill}

\subsection{Crossword Puzzle Class}

\begin{figure}[h!]
    \centering
    \subfloat[Original]{%
        \includegraphics[width=0.23\columnwidth]{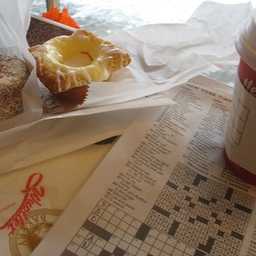}
    }
    \subfloat[No chequered]{%
        \includegraphics[width=0.23\columnwidth]{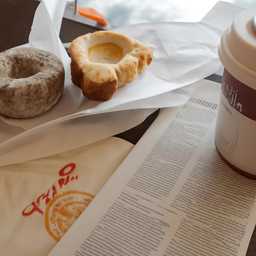}
    }
    \subfloat[No newspaper]{%
        \includegraphics[width=0.23\columnwidth]{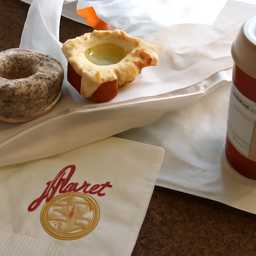}
    }
    \subfloat[No paper]{%
        \includegraphics[width=0.23\columnwidth]{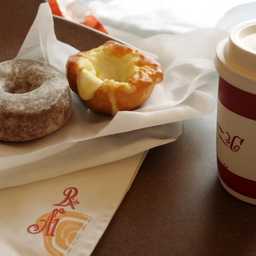}
    }
    \caption{Example of concept removal from a crossword puzzle class image, eliminating chequered, newspaper, and paper concepts.}
\end{figure}

\vspace*{\fill}

\subsection{Dalmatian Class}

\begin{figure}[h!]
    \centering
    \subfloat[Original]{%
        \includegraphics[width=0.25\columnwidth]{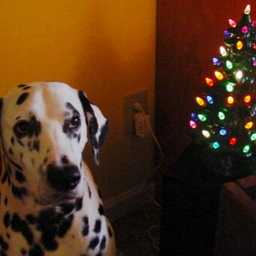}
    }
    \subfloat[No dotted]{%
        \includegraphics[width=0.25\columnwidth]{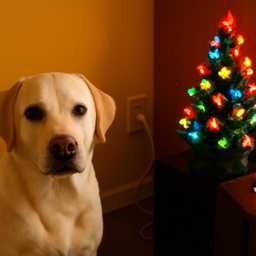}
    }
    \subfloat[No spotted]{%
        \includegraphics[width=0.25\columnwidth]{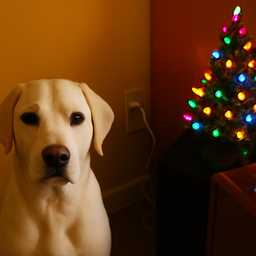}
    }
    \caption{Example of concept removal from a dalmatian class image, eliminating dotted and spotted concepts.}
\end{figure}

\vspace*{\fill}

\end{document}